%% file: main.tex
\documentclass{article}

    \PassOptionsToPackage{numbers, compress}{natbib}

\usepackage[preprint]{neurips_2026}

\input{preamble}

\usepackage[utf8]{inputenc} %
\usepackage[T1]{fontenc}    %
\usepackage{url}            %
\usepackage{booktabs}       %
\usepackage{amsfonts}       %
\usepackage{nicefrac}       %
\usepackage{microtype}      %
\usepackage{xcolor}         %

\definecolor{cvprblue}{rgb}{0.21,0.49,0.74}
\definecolor{lightcarminepink}{rgb}{0.9, 0.4, 0.38}
\usepackage[pagebackref,breaklinks,colorlinks,citecolor=cvprblue,linkcolor=lightcarminepink]{hyperref}

\title{Zero-shot 2D Grounding with Novel Affordance Types}

\author{%
  Haomeng Zhang\quad  Raymond A. Yeh\\
  Department of Computer Science, Purdue University\\
  \texttt{\{zhan5050, rayyeh\}@purdue.edu} \\
}

\begin{document}

\maketitle

\input{figs/new_teaser}

\input{sec/abstract}

\input{sec/intro}

\input{sec/rel}

\input{sec/preliminaries}

\input{sec/approach}

\input{sec/experiment}

\input{sec/conclusion}

\newpage
{
\small
\bibliographystyle{ieeenat_fullname}
\bibliography{main}
}

\clearpage
\input{supp_src/suppl}

\clearpage

\end{document}

%% file: preamble.tex
\newcommand*{\ShowNotes}{}
\usepackage{algorithm}
\usepackage{algorithmicx}
\usepackage{algpseudocode}
\input{macros}

\input{math_commands}

\usepackage{symbols}
\usepackage{pifont}
\usepackage{bbm}
\usepackage{float}
\usepackage{caption}
\usepackage{graphicx}
\usepackage{enumitem}
\usepackage{amssymb}
\usepackage{wrapfig}

\newcommand\ModelNamefree{\texttt{AffordAnything}} %
\newcommand\ModelNametrain{\texttt{\ModelNamefree{}+}}
\newcommand\benchmark{NAT}
\newcommand\agdbenchmark{AGD20K-\benchmark{}}
\newcommand\umdbenchmark{UMD-\benchmark{}}

\newcommand{\ccmark}{{\color{green!60!black}\ding{51}}}  %
\newcommand{\cxmark}{{\color{red}\ding{55}}}             %

\newcommand{\myparagraph}[1]{\vspace*{0pt}{\bf #1}}

%% file: macros.tex
\usepackage{color}
\usepackage{soul}
\usepackage{multirow}
\usepackage{xcolor}
\usepackage{colortbl}
\usepackage{wrapfig}

\definecolor{darkred}{rgb}{0.7,0.1,0.1}
\definecolor{darkgreen}{rgb}{0.1,0.7,0.1}
\definecolor{cyan}{rgb}{0.7,0.0,0.7}
\definecolor{dblue}{rgb}{0.2,0.2,0.8}
\definecolor{maroon}{rgb}{0.76,.13,.28}
\definecolor{burntorange}{rgb}{0.81,.33,0}
\definecolor{tealblue}{rgb}{0.212,0.459, 0.533}
\definecolor{myyellow}{rgb}{0.8627451 , 0.75294118, 0.20784314]}

\definecolor{mypink}{rgb}{0.93359375, 0.62109375, 0.83984375}

\definecolor{pp}{rgb}{0.43921569, 0.18823529, 0.62745098}
\definecolor{rr}{rgb}{0.5254902 , 0.00784314, 0.12941176}
\definecolor{bb}{rgb}{0.09019608, 0.23529412, 0.37647059}
\definecolor{yy}{rgb}{0.49803922, 0.3372549 , 0.0}
\definecolor{gg}{rgb}{0.02352941, 0.3372549 , 0.17647059}
\definecolor{mybrown}{rgb}{0.87058824, 0.56078431, 0.01960784}
\definecolor{myblue}{rgb}{0.3372549 , 0.70588235, 0.91372549}
\definecolor{mypurple}{rgb}{0.8, 0.47058824, 0.7372549 }
\definecolor{myorange}{rgb}{0.835, 0.368, 0}
\definecolor{mygreen}{rgb}{0.00784314, 0.61960784, 0.45098039}
\definecolor{mygt}{rgb}{0.0078125 , 0.57421875, 0.40625}
\definecolor{mysp}{rgb}{0.84765625, 0.515625  , 0.0234375}
\definecolor{mycitecolor}{rgb}{0,0.08,0.45}
\definecolor{mygr}{rgb}{0.12,0.6,0.12}
\definecolor{myoo}{rgb}{0.992,0.9176,0.9019}

\definecolor{myrr}{HTML}{AE031A}
\definecolor{mybb}{HTML}{0155B3}

\definecolor{myred}{rgb}{1, 0, 0}

\definecolor{lightgray}{gray}{0.9}

\definecolor{reblue}{rgb}{0.12,0.49,0.85}
\definecolor{reorange}{rgb}{0.835, 0.368, 0}
\definecolor{regreen}{rgb}{0.00784314, 0.61960784, 0.45098039}

\definecolor{trainfree}{rgb}{0.82, 0.88, 0.94}
\definecolor{trainable}{rgb}{1.0, 0.93, 0.78} 
\newcommand{\colortrainfree}{\rowcolor{trainfree}}
\newcommand{\colortrainable}{\rowcolor{trainable}}

\ifdefined\ShowNotes
  \newcommand{\colornote}[3]{{\color{#1}\bf{#2: #3}\normalfont}}
\else
  \newcommand{\colornote}[3]{}
\fi

\newcommand\checkclaim[1]{\textcolor{mysp}{#1}}
\renewcommand{\checkclaim}[1]{#1}

\newcommand\new[1]{\textcolor{dblue}{#1}}
\renewcommand{\new}[1]{#1}

%% file: math_commands.tex
\usepackage{amsmath,amsfonts,bm}

\def\1{\bm{1}}

\def\sp{space}

\def\va{{\bm{a}}}

\def\vc{{\bm{c}}}

\def\vf{{\bm{f}}}

\def\mI{{\bm{I}}}

\def\mK{{\bm{K}}}

\def\mM{{\bm{M}}}

\def\mQ{{\bm{Q}}}

\def\mV{{\bm{V}}}
\def\mW{{\bm{W}}}

\def\mY{{\bm{Y}}}

\DeclareMathAlphabet{\mathsfit}{\encodingdefault}{\sfdefault}{m}{sl}
\SetMathAlphabet{\mathsfit}{bold}{\encodingdefault}{\sfdefault}{bx}{n}

\def\gA{{\mathcal{A}}}

\def\gC{{\mathcal{C}}}

\def\gF{{\mathcal{F}}}

\def\gM{{\mathcal{M}}}

\def\gO{{\mathcal{O}}}

\def\sR{{\mathbb{R}}}

%% file: figs/new_teaser.tex
\begin{figure}[h]
\centering
\vspace{-.8cm}
\includegraphics[width=1\linewidth]{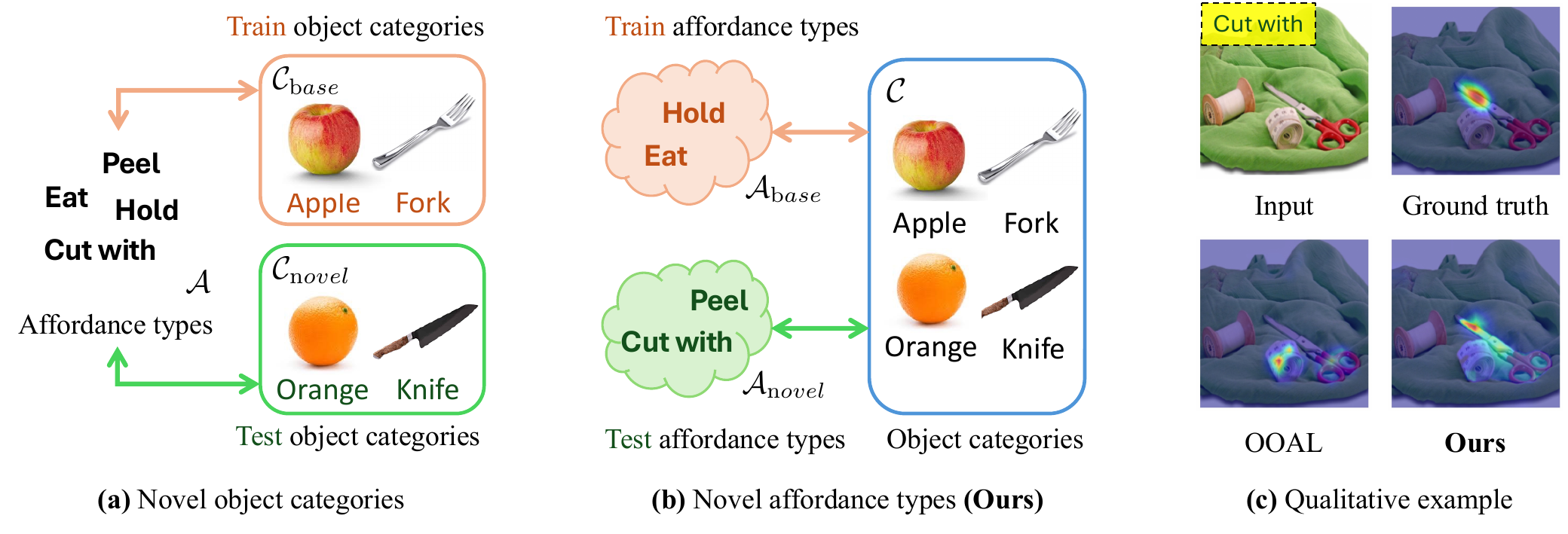}
\vspace{-0.4cm}
\caption{We propose to study \textbf{novel affordance types} in the task of zero-shot 2D grounding.
{\bf (a)} Prior works in 2D zero-shot affordance grounding focus on novel \textit{object categories}. {\bf (b)} In contrast, we propose to address models' generalization capability to novel \textit{affordance types}.
{\bf (c)} The state-of-the-art (SOTA) model OOAL~\cite{li:ooal:2024} does not generalize to affordance types that are \textit{unseen} during training, whereas our proposed method better predicts the interactive region for new affordance types.
}
\label{fig:new_teaser}
\vspace{-0.2cm}
\end{figure}

%% file: sec/abstract.tex
\begin{abstract}
2D affordance grounding aims to locate the region of an object that a human can interact with. Existing research focuses on recognizing affordance types seen during training and does not study models' ability to generalize to novel affordances, which is crucial for real-world applications. 
We propose the task of zero-shot 2D grounding with novel affordance types (NAT) and introduce the \benchmark{} benchmarks. 
We then propose \ModelNamefree{}, a training-free method that leverages segmentation cues, motivated by the strong correlation between affordance regions and object subparts.
To further improve performance, we develop \ModelNametrain{}, a trainable variant that learns to combine these cues. On the proposed \agdbenchmark{} benchmark,
our best model \ModelNametrain{} achieves a substantial improvement of 12.3\% (absolute) in IoU@0.4 over the SOTA affordance grounding method, OOAL.
\end{abstract}

%% file: sec/intro.tex
\section{Introduction}
\label{sec:intro}

Affordance understanding studies what actions are possible in a given environment~\cite{gibson2014ecological, visualafford}.
Unlike object recognition, which focuses on appearance, affordance requires reasoning about functionality,~\ie, what can be done with an object. One should be able to recognize that the handle of a knife is for grasping and identify the blade for cutting. This bridges perception and action, making affordance understanding essential for artificial intelligence. Grounding affordances is crucial for various applications, such as robot manipulation, action prediction, virtual reality, and scene understanding~\cite{ahn2022can, bahl2023affordances, geng2023rlafford, cheng2013affordances, vu2014predicting}.

Existing 2D affordance grounding methods~\cite{Learningluo, umd} mainly evaluate affordances seen during training.
Zero-shot benchmarks have been proposed to test \textit{object categories} not present in the training set.
Recent studies~\cite{li:ooal:2024, jang2024intra, cuttano2024does} leverage pretrained VLMs to improve the grounding performance for novel {\it object} classes. However, the generalization to novel {\it affordance} types (NAT)
has not been fully studied. See~\figref{fig:new_teaser} (a) vs. (b) for a comparison. In (a), the test objects are new, but the affordance types overlap with training. In the NAT setting (b), the objects are seen during training, but the affordance type (\eg, cut) is new.
Such generalization to novel affordances is necessary in real-world applications, where agents may receive open-ended instructions and must localize the corresponding actionable regions without having seen those affordance types during training.

To the best of our knowledge, only a few works~\cite{cuttano2024does,jang2024intra} provide \textit{qualitative} demonstrations, and \textbf{none}~\cite{Learningluo, zhai2022one, li2023locate, xu2024weakly, li:ooal:2024, rai2024strategies, qian2024affordancellm, cuttano2024does, tong2024oval, ju2024robo, jang2024intra, xuweakly, li2024learning, moon2025selective} \textbf{have systematically evaluated} generalization to novel affordance types.
When evaluating the SOTA model~\cite{li:ooal:2024}, we find that it does not reliably generalize to novel affordance types (see~\figref{fig:new_teaser} (c)).

This motivates us to analyze and develop models that can effectively handle zero-shot grounding of novel affordances.
We explicitly split affordance types in existing datasets into disjoint training and testing sets.
This allows us to recast existing affordance datasets, \eg,
AGD20K~\cite{Learningluo} and UMD~\cite{umd}, in the NAT setting, referred to as \agdbenchmark{} and \umdbenchmark{}.
We then propose \ModelNamefree{}, a training-free approach that leverages pretrained VLMs to extract affordance regions from \textit{segmentation cues} based on object subparts.
We further introduce \ModelNametrain{}, which incorporates a trainable fusion module to learn how to combine these cues to improve performance further.
To maintain the generalization of the local segmentation cues, \ModelNametrain{} only learns lightweight fusion weights, reducing overfitting to the training samples.

Empirically, we compare our models against several SOTA affordance grounding and open-vocabulary segmentation models~\cite{li:ooal:2024, bousselham2024grounding, wang2024sclip, lan2024proxyclip, lan2024clearclip}. %
Experiments show that existing methods struggle on the \benchmark{} benchmarks. This indicates that \checkclaim{the proposed problem setting is challenging and that VLMs alone are not sufficient for this task}.
Our proposed \ModelNamefree{} and \ModelNametrain{} consistently outperform all baselines, demonstrating the effectiveness of our approach. 
On the \agdbenchmark{}, \ModelNamefree{} achieves a 13.1\% absolute improvement in IoU@0.4 over the SOTA training-free open-vocabulary segmentation models.
With one-shot training, \ModelNametrain{} outperforms the SOTA affordance grounding method by 12.3\% (absolute) in IoU@0.4.

{\bf\noindent Our main contributions are as follows:}
\vspace{-3pt}
\begin{itemize}[topsep=0pt, leftmargin=12pt]
    \setlength{\itemsep}{0.0pt}
    \setlength{\parskip}{2.5pt}
    \item We propose to study the task of zero-shot 2D grounding with \textit{novel affordance types}
    and establish \benchmark{} benchmarks to evaluate SOTA baselines.
    \item To tackle this task, we develop \textbf{\ModelNamefree{}}, a training-free pipeline for affordance grounding, and \textbf{\ModelNametrain{}}, a trainable version that further improves performance.
    \item Experiments on \agdbenchmark{} and \umdbenchmark{} benchmarks validate the effectiveness of our framework. Our approach consistently outperforms SOTA baselines in grounding novel affordance types. Code will be made available.
\end{itemize}

%% file: sec/rel.tex
\section{Related Work}
\label{sec:rel}

{\bf\noindent 2D affordance grounding} involves identifying the regions of an object with which a human can interact. Early approaches rely on CNNs to predict affordance regions~\cite {chuang2018learning, do2018affordancenet, myers2015affordance}.
Other works explore learning affordances from videos~\cite{chen2023affordance, fang2018demo2vec, luo2023learning, nagarajan2019grounded, koppula2013learning, heidinger20252handedafforde, li2024learning}, leveraging temporal interactions and motion cues to improve affordance grounding.
Due to limited dense annotations, researchers have studied weakly supervised methods~\cite{cornia2016deep, huang2018predicting, kummerer2016deepgaze, luo2023learning, nagarajan2019grounded, pan2017salgan, sawatzky2017weakly, nagarajan2020learning, xuweakly, moon2025selective, huang2025resource, tang2025closed}, which learn to predict affordance regions from sparse annotations, such as keypoints~\cite{cui2023strap, sawatzky2017adaptive} or image-level labels~\cite{li2023locate, Learningluo, xu2024weakly}.
However, all of these methods assume a closed set of affordance types during training and testing, and thus cannot adapt or generalize well to novel affordances.

Recently, foundation models have been used to improve affordance grounding~\cite{jiang2025affordancesam}. %
CLIP-based methods~\cite{nguyen2023open, rashid2023language, ju2024robo} use visual/textual embeddings from contrastive learning to associate visual representations with textual descriptions.
DINO-based backbones~\cite{caron2021emerging, oquab2023dinov2} are also widely adopted~\cite{li2023locate, rashid2023language, jang2024intra, hadjivelichkov2023one, li:ooal:2024, lu2025geal, jia2025one} due to their more comprehensive visual representations.
Another direction~\cite{mees2023grounding, huang2023voxposer, mirjalili2023lan, song2023learning, qian2024affordancellm, chen2024worldafford, ELLMER, yu2025seqafford, wei20253daffordsplat, wu2025ragnet} incorporates LLMs~\cite{brown2020language, achiam2023gpt, chu20253d} to inject commonsense reasoning into affordance grounding.

While these methods improve affordance grounding, %
they primarily focus on the generalization ability to new object categories. %
Despite the strong representations provided by the foundation models, current approaches~\cite{li:ooal:2024} still struggle to generalize to \textit{novel affordances} due to the %
predefined/fixed, closed set of affordance types. 
In contrast, we aim to extend the model's generalization capability to novel affordances. %
This is done by leveraging affordance-independent local segmentation cues.
In the Appendix~\tabref{tab:rel_work}, we provide a detailed tabularized comparison of prior works.

\input{figs/pipeline}
\myparagraph{Zero-shot learning} aims to recognize
objects or concepts not seen during training~\cite{wu2024towards}.
Substantial progress has been made in zero-shot dense prediction, particularly in 2D semantic segmentation~\cite{ding2022decoupling, zhou2022extract, li2022adapting, zhou2023zegclip, xu2023side, xie2024sed, lan2024proxyclip, lan2024clearclip}.
The typical zero-shot setting in 2D affordance grounding focuses on novel object categories, where the model is evaluated on images containing objects whose categories were not present in the training set~\cite{Learningluo, wei2025afforddexgrasp}.
Recently, the 3D community has introduced the setting that tests zero-shot capabilities for new affordance types. %
However, these settings either restrict the affordance types to only synonyms of training affordances~\cite{van2024open, nguyen2023open}, or require additional demo images as input~\cite{shao2024great}.
Different from previous work, this work introduces a \textit{stricter} zero-shot setting on \textit{novel affordance types} in 2D, and we do not rely on additional auxiliary inputs.

%% file: figs/pipeline.tex
\begin{figure}[t]
\centering
\includegraphics[width=\linewidth]{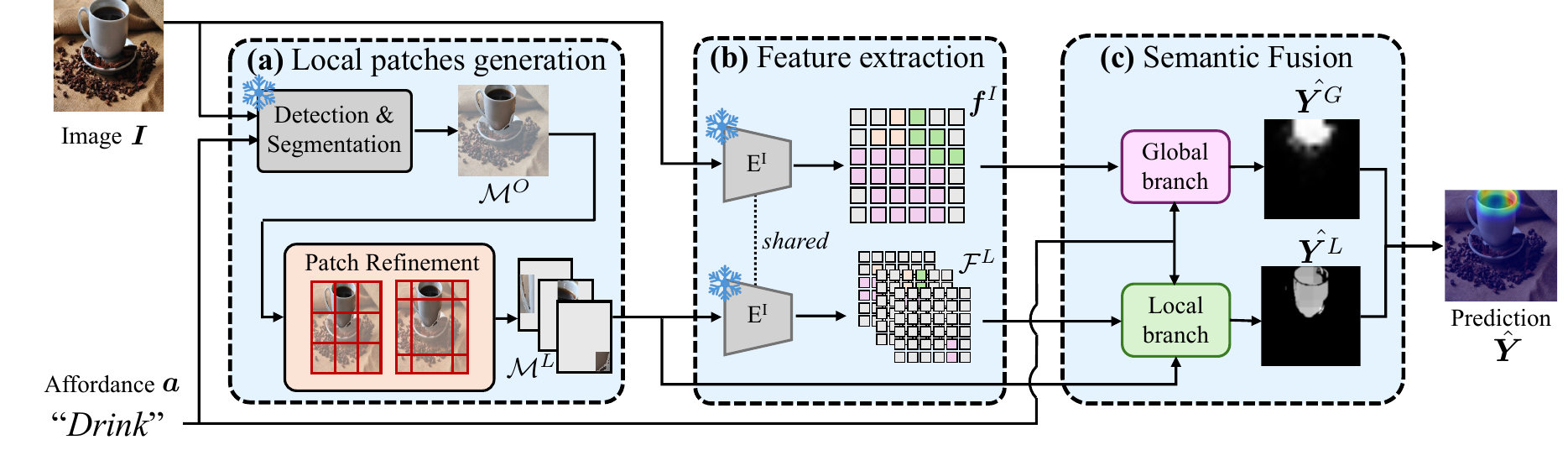}
\vspace{-0.5cm}
\caption{\textbf{Illustration of \ModelNamefree{}} (\secref{sec:trainingfree}).
\ModelNamefree{} first generates local patch candidates through the (a) detection-and-segmentation backbone and the patch refinement module. Next, 
(b) the visual and text features are extracted and (c) fused in the semantic fusion module in a multi-branch manner.
The frozen text encoder $\operatorname{E^T}$ is omitted in the visualization for clarity.
}
\vspace{-0.3cm}
\label{fig:model}
\end{figure}

%% file: sec/preliminaries.tex
\section{Background}
\label{sec:pre}

{\bf \noindent 2D affordance grounding.} Given an image $\mI \in \sR^{H\times W\times 3}$ 
and an affordance label $\va$ expressed in natural language (\eg, `drink'), the goal of 2D affordance grounding is to predict a pixel-level affordance mask $\mM \in \sR^{H\times W}$ 
that localizes the corresponding actionable region.
The predicted mask $\mM$ can either be a probability heatmap or a binary mask.

\myparagraph{Existing zero-shot setting.} The ``zero-shot'' setting in existing benchmarks~\citep{Learningluo, umd} evaluates a model’s ability to generalize to novel {\it object categories}.
All available object classes $\gC$ are divided into a disjoint base class set $\gC_{\text{base}}$ and a novel class set $\gC_{\text{novel}}$. %
During evaluation, the model is tested on images whose object class label $\vc_\text{{test}} \in \gC_{\text{novel}}$, while it was trained only on images with the object class label $\vc_\text{{train}} \in \gC_{\text{base}}$.
In this setting, the affordance input
$\va$ for both training and testing comes from the {\it same pre-defined affordance set} $\gA$. 
Note that \emph{zero-shot} means that no annotated training pairs whose class labels belong to $\gC_{\text{novel}}$ are used during training. This setting does not prohibit the use of pre-trained backbones, following the convention in prior works,~\eg, OOAL~\cite{li:ooal:2024}. %

%% file: sec/approach.tex
\section{Approach}
\label{sec:approach}

We aim to develop a model for affordance grounding that effectively handles novel affordance types in a zero-shot manner.
First, we introduce the problem formulation (\secref{sec:problem}).
Next, we describe the proposed training-free method \ModelNamefree{} (\secref{sec:trainingfree}), which uses a pretrained VLM to predict the region for any image with respect to any affordance.
Finally, we propose \ModelNametrain{} (\secref{sec:trainable}), a trainable fusion module to adapt \ModelNamefree{} to specific data domains.

\subsection{Problem formulation}
\label{sec:problem}
\input{figs/motivation}
Given an image $\mI \in \sR^{H\times W\times 3}$ 
and an affordance type label $\va$ expressed in natural language, \eg, `drink', the task of 2D affordance grounding aims at predicting the pixel-level affordance mask $\mY \in \sR^{H\times W}$ 
indicating the interactive region.
To evaluate the model's generalization ability to \textit{novel affordances}, %
we divide all available affordance classes $\gA$ into two disjoint sets: the base set $\gA_{\text{base}}$ and the novel set $\gA_{\text{novel}}$. %
The model will be trained on image-affordance pairs where $\va_\text{{train}} \in 
\gA_{\text{base}}$, but tested on pairs where $\va_\text{{test}} \in \gA_{\text{novel}}$.
\figref{fig:new_teaser} illustrates and compares the task with existing benchmarks.

\myparagraph{Challenges.} 
Training-based affordance grounding methods assume the same fixed set of affordance types for both training and testing.
As a result, these models tend to learn only the semantics associated with those specific affordances, limiting their generalization capability to novel affordances.
To mitigate this issue, we build models that leverage a pretrained VLM to generalize. This is done by having either no training or only minimal fine-tuning on limited training data.

\subsection{AffordAnything}
\label{sec:trainingfree}
We propose a training-free method, \ModelNamefree{}, based on segmentation cues/features from pretrained VLMs to predict the mask for any affordance. This paradigm involves three components:
{\bf (i)} a local patches generation process to produce potential affordance regions; 
{\bf (ii)} a feature extraction module over image, patch, and text features; 
{\bf (iii)} a semantic fusion module that combines the visual and text features to predict the affordance mask in a multi-branch manner. 
An overview of \ModelNamefree{} is illustrated in~\figref{fig:model}. For readability, we defer many of the implementation details to the Appendix~\secref{sec:implementation_details}.

\input{figs/localmask}
\textbf{Generating local patches.} 
Inspired by the observation that affordance regions strongly correlate with specific sub-parts of the object (see \figref{fig:motivation}), we aim to detect the object and extract a set of local region candidates that potentially contain the affordance.
We adopt an open-vocabulary detection and segmentation backbone Grounded-SAM~\cite{ren2024grounded}
to detect $N$ objects $\gO = \{O_1, \ldots, O_N \}$ along with their corresponding masks $\gM^O = \{\mM^O_1, \ldots, \mM^O_N \}$.
Since explicit object category labels are not available, we query the backbone directly with the affordance label $\va$ to obtain associated masks.
Benefiting from large-scale pretraining on diverse text captions~\cite{radford2021learning}, the backbone can detect objects relevant to the given affordance query:
\bea
\gO, \gM^O = \operatorname{Detection-and-Segmentation}(\mI, \va).
\eea

Next, affordance regions often align with specific object parts or boundaries.
For example, grasping a bottle involves interacting with its central body, while opening a book requires locating its edge. Hence, we capture these fine-grained regions by further decomposing each detected object mask $\mM^O_n$ for object $O_n$ into two sets of spatial patches: evenly-focused patches $\gM^E_n$ and border-focused patches $\gM^B_n$.
This decomposition is achieved through a patch refinement process, formulated as
\bea
\gM^E_n, \gM^B_n = \operatorname{Patch-Refinement}(\mM^O_n)
\eea
and illustrated in~\figref{fig:localmask}.
Each object mask $\mM^O_n$ is refined into $K$ local patches 
$\{\mM^{L}_{n1}, \ldots, \mM^{L}_{nK}\}$
based on the relative spatial position.
We use the notation $\gM^L_n$ to subsume all local patches,~\ie, the original object mask $\mM^O_n$, evenly-focused patches $\gM^E_n$, and border-focused patches $\gM^B_n$. 
The final local patch candidates
$
\gM^{L} = \{\gM^L_1, \ldots, \gM^L_N \}
$
are the union of all local patches $\gM^L_n$.

\myparagraph{Feature extraction.}
To utilize the generalization of a pre-trained VLM, we use a frozen 
image encoder $\operatorname{E^I}$ and text encoder $\operatorname{E^T}$ from CLIP~\cite{radford2021learning} to extract visual/text features. Visual features are extracted at two hierarchies: an image-level feature $\vf^I$ from $\mI$ and patch-level features $\vf^L_{nk}$ corresponding to each patch indicated by the mask $\mM^{L}_{nk}$ of the $n$-th object and the $k$-th local patch. The text feature $\vf^t$ is computed from the input affordance type $\va$. In other words, 
\bea
\vf^I = \operatorname{E^I}(\mI), \quad \vf^L_{nk} = \operatorname{E^I}(\mI \odot \mM^{L}_{nk}), \quad \vf^t = \operatorname{E^T}(\va),
\eea
where $\odot$ denotes element-wise multiplication to generate a masked image.
In summary, the patch features form a set 
\bea
\gF^{L} = \{\gF^L_1, \ldots, \gF^L_N \}, \text{ where } \gF^L_n = \{\vf^{L}_{n1}, \ldots, \vf^{L}_{nK}\}.
\eea

\myparagraph{Semantic fusion.}
We aggregate all the extracted features using the fusion module to make a final prediction $\hat{\mY}$ in a multi-branch manner.
To capture both the image-level and local region details, we propose to generate the probability mask from a global branch and a local branch separately and then fuse them:
\bea
\label{eqn:branch_weight}
\hat{\mY} = \alpha \cdot \hat{\mY}^G + (1 - \alpha) \cdot \hat{\mY}^L.
\eea
Here, $\alpha \in [0, 1]$ is the branch weight and $\hat{\mY}^{G/L}$ denotes the prediction from the global/local branch.

In more detail, the local branch prediction is obtained by aggregating object-level predictions via element-wise maximization 
\bea\label{eqn:multi_branch_l}
\hat{\mY}^L \triangleq \operatorname{Local\text{-}branch}(\gM^L, \gF^L, \vf^t)
= \bigvee_{n=1}^{N} \hat{\mY}^L_n,
\eea
where $\bigvee$ denotes the element-wise maximum across spatial locations. Each local prediction $\hat{\mY}^L_n$ for object $O_n$ is modeled as an attention of associated local patches:
\bea\label{eqn:mask_weights}
\hat{\mY}^L_n = \sum_{k=1}^{K} w_{nk} \mM_{nk}, \text{ where }~ w_{nk} =
\frac{\exp\big(\operatorname{cos}(\vf^L_{nk}, \vf^t)\big)}
{\sum_{j=1}^{K} \exp\big(\operatorname{cos}(\vf^L_{nj}, \vf^t)\big)}
\eea
are computed from the cosine similarity of the patch feature $\vf^L_{nk}$ and the text feature $\vf^t$.

For the global branch, we model the $\operatorname{Global-branch(\cdot)}$ as the normalized attention map between the image feature $\vf^I$ and the text feature $\vf^t$:
\bea
\label{eqn:multi_branch_g}
\hat{\mY}^G \triangleq
\operatorname{Global-branch}(\vf^I, \vf^t) = \operatorname{\sigma}\left(\frac{\vf^t(\vf^I)^\top}{\sqrt{d}}\right),
\eea
where $\sigma(\cdot)$ is the sigmoid function and $d$ is a scaling factor.

\subsection{AffordAnything+}
\label{sec:trainable}
To further improve \ModelNamefree{}'s performance, we propose \ModelNametrain{} to integrate a trainable fusion module that aggregates and preserves the training-free affordance-independent local segmentation cues.
As shown in~\figref{fig:fusion}, the trainable fusion module contains (i) a branch weight predictor, (ii) a global branch decoder, and (iii) a local weight predictor.
Specifically, we learn the branch weight $\alpha$ in \equref{eqn:branch_weight}, and design learnable $\operatorname{Local-branch(\cdot)}$ and $\operatorname{Global-branch(\cdot)}$ in \equref{eqn:multi_branch_l} and \equref{eqn:multi_branch_g}.
The trainable fusion module follows the multi-branch design and learns refined image-level semantics and local region details.

\myparagraph{Branch weight predictor.}
To learn the relationship between local branch prediction $\hat{\mY}^L$ and global branch prediction $\hat{\mY}^G$, we model the learnable branch weight $\alpha$ in~\equref{eqn:branch_weight} as a linear projection of the image feature $\vf^I$ and the text feature $\vf^t$, followed by a sigmoid activation:
\bea
\alpha^{*} = \sigma\!\left(\operatorname{Linear^W}(\vf^{I} \oplus \vf^{t})\right),
\eea
where $\oplus$ denotes concatenation and $\sigma(\cdot)$ is the sigmoid function.
This enables the model to dynamically adjust the relationship between global prediction and local prediction based on the image input and the affordance label.

\myparagraph{Local weight predictor.}
To preserve training-free affordance-independent local segmentation cues for generalization, we only dynamically learn the patch weights $w_{nk}$ for each local patch candidate in $\operatorname{Local-branch(\cdot)}$.
We model an additional similarity bias $\Delta s_{nk}$ as a linear function of each patch feature $\vf^{L}_{nk}$ and the text feature $\vf^{t}$:
\bea
\Delta s_{nk} = \operatorname{Linear^S}(\vf^{L}_{nk} \oplus \vf^{t}).
\eea
The local patch weight in \equref{eqn:mask_weights} is now learned as:
\bea
w^{*}_{nk}=\frac{\operatorname{exp}(\operatorname{cos}(\vf^L_{nk}, \vf^t) + \Delta s_{nk})}{\sum_{j=1}^{K}\operatorname{exp}(\operatorname{cos}(\vf^L_{nj}, \vf^t) + \Delta s_{nj})}.
\eea

\input{figs/fusion}
\textbf{Global branch decoder.}
Given the image feature $\vf^I$ and text feature $\vf^t$, the global branch decoder predicts the global branch mask $\hat{\mY}^G$.
The global branch decoder is a stack of transformer layers and follows the standard cross-attention mechanism~\cite{vaswani2017attention} consisting of queries, keys, and values. Given the image feature $\vf^I$ and the text feature $\vf^t$, the queries $\mQ$, keys $\mK$, and values $\mV$ are computed as: $\mQ = \vf^t\mW_Q, \mK = \vf^I\mW_K, \mV = \vf^I\mW_V$
with linear projections $\mW_{Q/V/K}$. 
Following \citet{li:ooal:2024}, we refine the text embedding by retrieving relevant visual information: $\tilde{\vf^{t}} = \operatorname{Softmax}(\frac{\mQ\mK^\top}{\sqrt{d}})\mV$.
Finally, we substitute $\vf^{t}$ in~\equref{eqn:multi_branch_g} with the updated text feature $\tilde{\vf^{t}}$ and obtain the learnable global branch prediction:
$
\operatorname{Global-branch^{*}}(\vf^I, \vf^t) = \operatorname{Sigmoid}\left(\frac{\tilde{\vf^{t}}(\vf^I)^\top}{\sqrt{d}}\right).
$

%% file: figs/motivation.tex
\begin{wrapfigure}[15]{r}{0.5\linewidth}
\centering
\vspace{-0.2cm}
\includegraphics[width=1\linewidth]{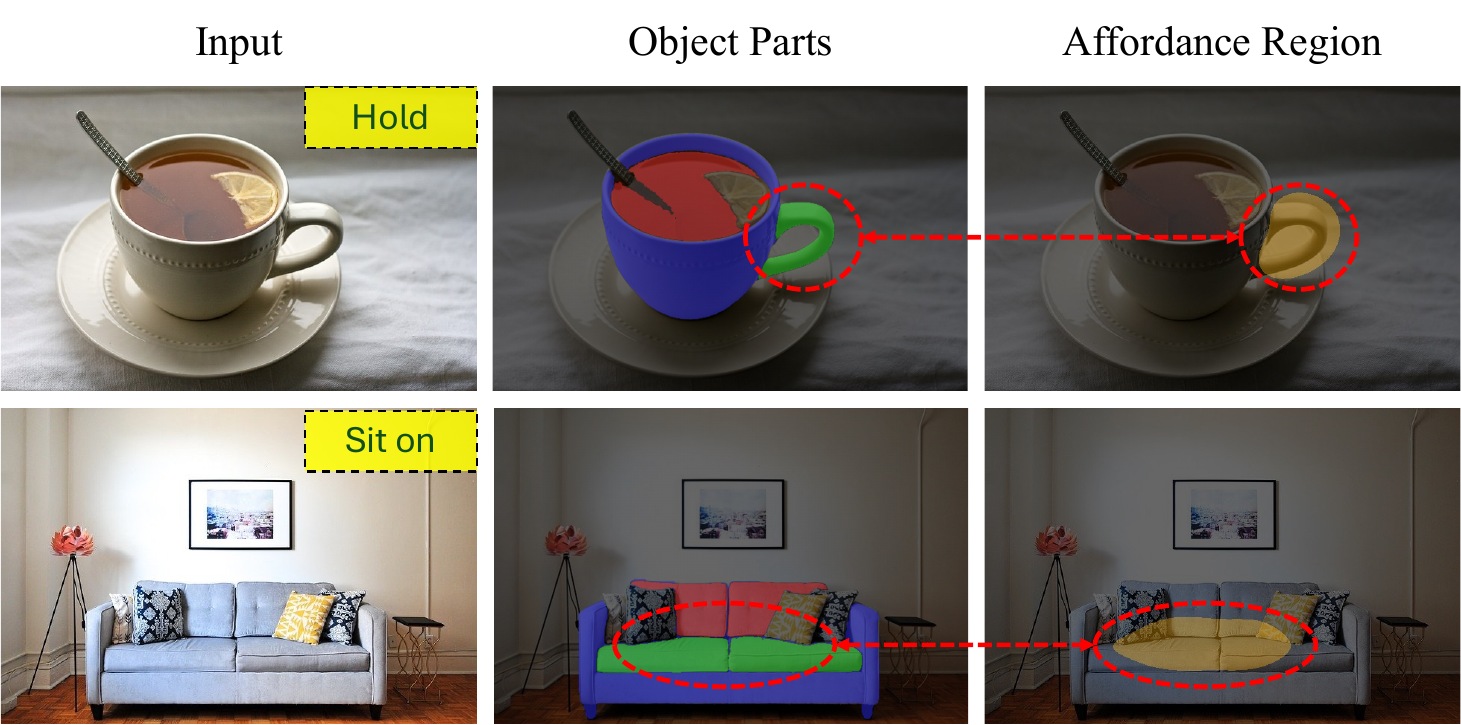}
\caption{\textbf{Motivation for local patches generation}.
The \textcolor{yy}{affordance region} typically
corresponds to
a \textcolor{mygr}{sub-region} within the object mask.
}
\label{fig:motivation}
\end{wrapfigure}

%% file: figs/localmask.tex
\begin{wrapfigure}[20]{r}{0.4\linewidth}
\centering
\vspace{-1cm}
\includegraphics[width=1\linewidth]{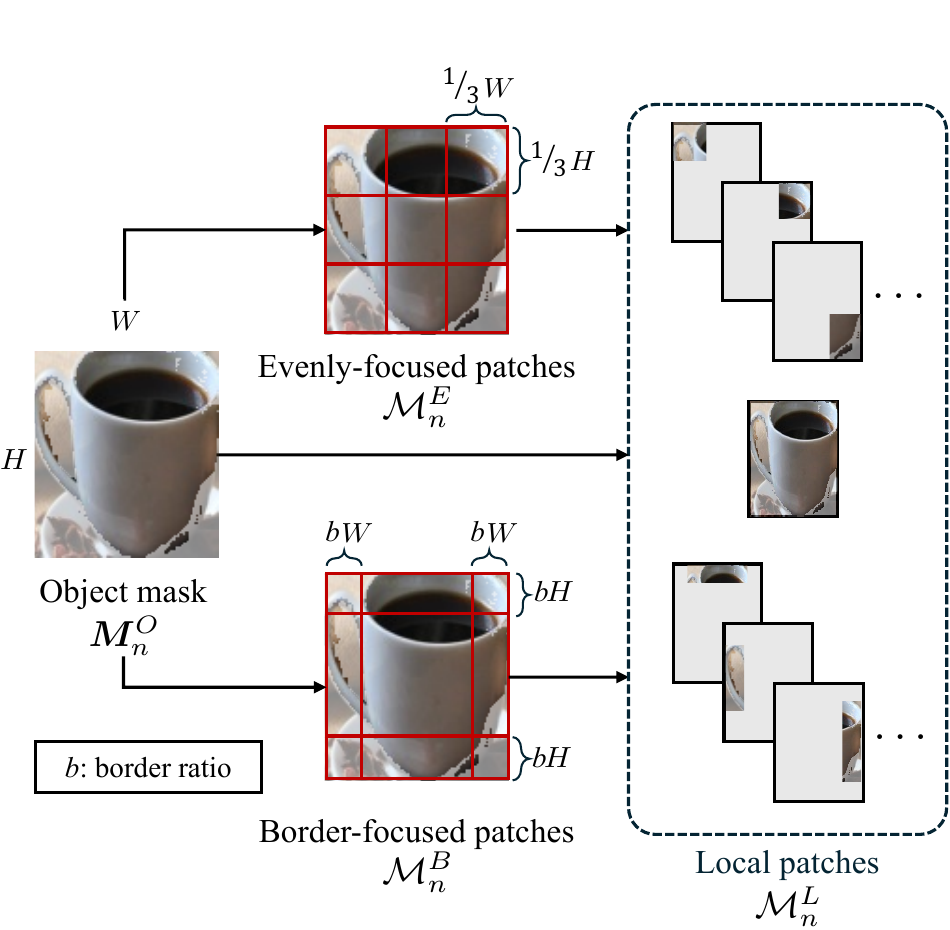}
\caption{\textbf{Patch refinement process.} 
For each object mask, we generate evenly-focused patches $\gM_n^{E}$ 
and border-focused patches $\gM_n^{B}$ 
to capture the local details based on the spatial relations.
}
\label{fig:localmask}
\end{wrapfigure}

%% file: figs/fusion.tex
\begin{wrapfigure}[17]{r}{0.48\linewidth}
\vspace{-0.3cm}
\centering
\includegraphics[width=\linewidth]{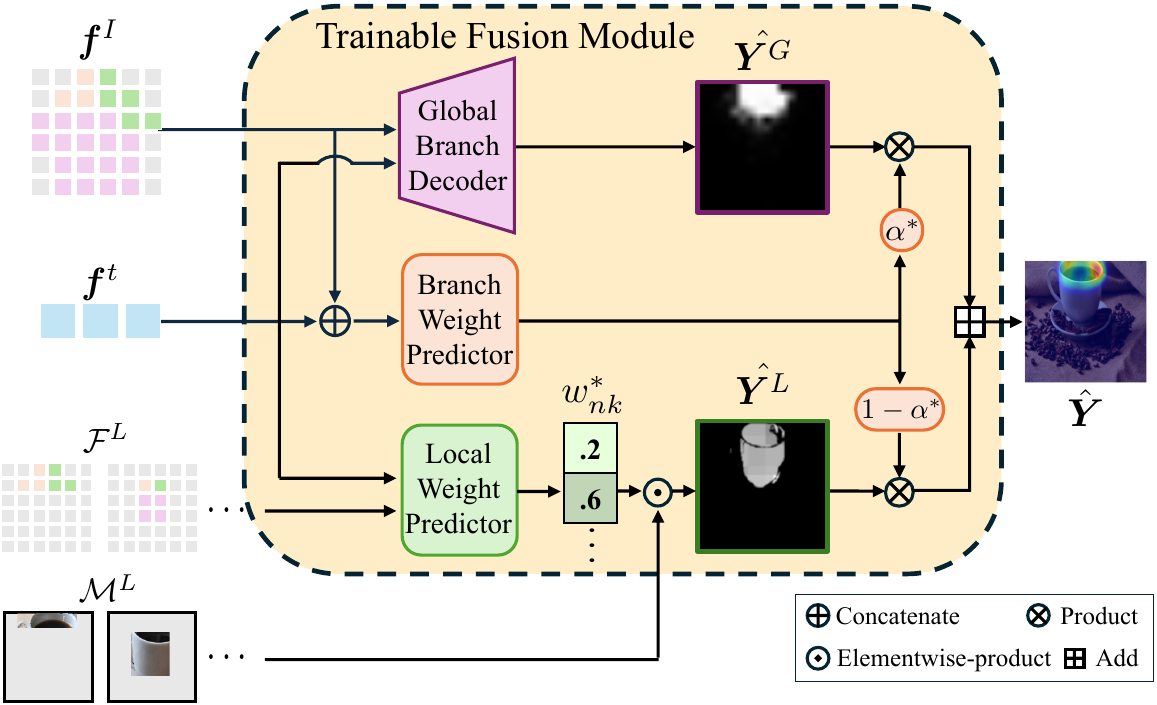}
\caption{\textbf{Trainable fusion module} in \ModelNametrain{} (\secref{sec:trainable}).
The global branch prediction $\hat{\mY}^G$ and local branch prediction $\hat{\mY}^L$ are weighted by the learned branch weight $\alpha^*$. %
}
\label{fig:fusion}
\end{wrapfigure}

%% file: sec/experiment.tex
\section{Experiments}
\label{sec:experiment}

First, we establish the \benchmark{} benchmarks for our proposed task of zero-shot 2D grounding of novel affordance types (\secref{sec:exp_benchmark}).
Next, we conduct experiments on our \agdbenchmark{} and \umdbenchmark{} benchmarks. 
We compare our models with other open-source 2D affordance grounding models and open-vocabulary segmentation models in both training-free (\secref{sec:exp_trainingfree}) and trainable (\secref{sec:exp_trainable}) settings. 
Finally, we ablate the proposed designs (\secref{sec:exp_ablation}).

\subsection{NAT Benchmark settings}
\label{sec:exp_benchmark}

{\bf \noindent Datasets.} 
We use AGD20K~\cite{Learningluo} and UMD part affordance~\cite{umd} datasets to establish the \agdbenchmark{} and \umdbenchmark{} benchmarks by re-splitting them following our NAT formulation in~\secref{sec:problem}.
Our splits are divided into train, val, and test sets, where the affordance types in the train, val, and test sets are mutually exclusive, with no overlap across splits. See Appendix~\secref{sec:datasplit} for details.

\myparagraph{Baselines.} We consider several open-source affordance grounding and open-vocabulary segmentation methods.
For 2D affordance models, we include OOAL~\cite{li:ooal:2024}, which can be adapted to unseen affordances.
For open-vocabulary segmentation, we include CLIP-based training-free models: GEM~\cite{bousselham2024grounding}, SCLIP~\cite{wang2024sclip}, ClearCLIP~\cite{lan2024clearclip}, ProxyCLIP~\cite{lan2024proxyclip}.
For a fair comparison, our training-free variant, \ModelNamefree{}, uses the same image backbone (CLIP~\cite{radford2021learning}) as these segmentation baselines.
In the trainable setting, \ModelNametrain{} adopts DINO~\cite{oquab2023dinov2} as the image encoder, consistent with the affordance grounding baseline~\cite{li:ooal:2024}.
We also compare against multi-modal large language models (MLLMs), including \texttt{GPT-5.2}~\cite{openai2026gpt52} and an open-source 2D VLM \texttt{Florence2}~\cite{xiao2024florence} in~\secref{sec:exp_trainingfree}. The most related LLM-based methods for 2D affordance reasoning are AffordanceLLM~\cite{qian2024affordancellm} and WorldAfford~\cite{chen2024worldafford}. However, they do not provide open-source implementations.
Implementation details are provided in~\secref{sec:baseline_models}.

\myparagraph{Training settings.} For the training-free setup, we directly evaluate our \ModelNamefree{} and other training-free open-vocabulary segmentation models on the test set of \agdbenchmark{} and \umdbenchmark{}. For the trainable setting, our \ModelNametrain{} and other baseline models are fine-tuned on \agdbenchmark{} and \umdbenchmark{} training sets.
This setup follows prior work for a fair comparison under limited supervision. %

\myparagraph{Evaluation metrics.} Following prior work~\cite{li:ooal:2024}, we report Intersection-over-Union at the threshold of 0.4 (IoU@0.4) for the UMD dataset.
For the AGD20K dataset, we report Kullback-Leibler Divergence (KLD), Similarity (SIM), and Normalized Scanpath Saliency (NSS). We also report IoU@0.4 to evaluate training-free baselines.

\subsection{Results for training-free setting}
\label{sec:exp_trainingfree}
\myparagraph{Comparison with segmentation models.}
We compare our model and the SOTA training-free open-vocabulary segmentation models on \agdbenchmark{} and \umdbenchmark{}, as presented in~\tabref{tab:main_training_free}. 
Our \ModelNamefree{} achieves an absolute increase of 13.1\% and 23.9\%, respectively, compared to the second-best models. 
We observe that our proposed \ModelNamefree{} is robust across different datasets, \checkclaim{which validates our idea that local segmentation cues are helpful for affordance grounding}.

\myparagraph{Comparison with MLLMs.}
We evaluate \texttt{GPT-5.2} and \texttt{Florence-2} on the affordance grounding task. Since \texttt{GPT-5.2} cannot directly perform dense prediction, we prompt it to predict the affordance region by outputting bounding box coordinates. The prompts are provided in~\secref{sec:baseline_models} in the supplement. As shown in~\figref{fig:gpt5}, \texttt{GPT-5.2} cannot reliably locate the affordance regions.
We further compare with \texttt{Florence-2} \cite{xiao2024florence}, a unified multimodal foundation model supporting pixel-level segmentation without task-specific fine-tuning.
Although pre-trained on the large-scale FLD-5B dataset \cite{xiao2024florence}, \texttt{Florence-2} achieves only an IoU@0.4 of 14.1 on \agdbenchmark{}, substantially lower than our \ModelNamefree{} (21.3).

These results indicate training-free open-vocabulary segmentation models and general-purpose 2D VLMs cannot be \emph{directly} applied to the affordance grounding task, \checkclaim{highlighting the inherent gap between the segmentation task and the affordance grounding task}.
Our patch refinement and multi-branch semantic fusion modules effectively mitigate this gap. %

\begin{table}[t]
    \begin{minipage}[t]{0.37\linewidth}
        \input{tabs/main_training_free}
    \end{minipage}
    \hfill
    \begin{minipage}[t]{0.6\linewidth}
        \input{tabs/main_trained}
    \end{minipage}
\vspace{-0.3cm}
\end{table}

\subsection{Results for trainable setting}
\label{sec:exp_trainable}

{\bf \noindent Comparison with SOTA.} 
In~\tabref{tab:main_trained}, we report SOTA affordance grounding methods fine-tuned on the \agdbenchmark{} and \umdbenchmark{} benchmarks.
We observe that for the \agdbenchmark{} benchmark, where the data annotation is probability heatmaps, expert affordance grounding models outperform the fine-tuned segmentation models over the distribution-oriented metrics (KLD, SIM and NSS). 
However, for the IoU@0.4 metric on the \agdbenchmark{} and \umdbenchmark{} benchmarks, fine-tuned open-vocabulary segmentation models can also achieve competitive performance.
This suggests that \checkclaim{such segmentation models could localize the high-confidence target regions and have a strong potential for generalizing to the affordance grounding}.

\myparagraph{Ablation on different pre-trained backbones.}
To verify how much the performance gain is from the more modern backbone, 
we conduct an additional variant of our \ModelNametrain{} trained with the CLIP backbone for a direct comparison across architectures in Appendix~\tabref{tab:supp_ablation_backbone}. Across different image encoders, our method {\bf consistently outperforms OOAL.}
Next, as our pipeline incorporates Grounded-SAM, we further provide SAM features to OOAL, which results in a KLD of 1.653. This is comparable to using DINO features (1.617), yet, still substantially worse than our model (1.374) as shown in~\tabref{tab:main_trained}. From these results, we conclude that the improvement primarily comes from our proposed design rather than the backbones.

\input{tabs/add_datasplit_results}
\textbf{Robustness across data splits.}
To further evaluate the robustness of our model, we analyze performance under different data splits on \agdbenchmark{}.
Specifically, we consider two additional configurations:
(i) \emph{Additional split 1}: We switch the validation and test sets of our \agdbenchmark{}, \ie, the validation and test partitions are swapped (details in~\tabref{tab:supp_datasplits});
(ii) \emph{Additional split 2}: We construct a new training set that covers a different subset of affordance categories while maintaining a comparable sample size to the original \agdbenchmark{} setup (see~\tabref{tab:supp_add_datasplits}).
The results for both additional splits are reported in~\tabref{tab:add_datasplit_results}.
We observe that our proposed \ModelNametrain{} remains robust across different data splits, \ie, it generalizes consistently to novel affordance types even under varying configurations.

\input{figs/qualitative}

\myparagraph{Qualitative results.}
In~\figref{fig:qualitative}, we visualize the grounded affordance for \agdbenchmark{}. 
We find that \checkclaim{the expert affordance grounding model, OOAL, fails to localize the corresponding affordance regions in most cases}.
This is mainly caused by its learnable text prompt design, which overfits to the training affordance types. 
On the other hand, fine-tuned open-vocabulary segmentation baselines better capture the object-level semantic patterns in some cases but fail to localize the specific subpart associated with the affordance input.
This suggests that, even with open-vocabulary object recognition capabilities, \checkclaim{directly applying large-scale pre-trained VLMs is insufficient for the affordance type generalization}, highlighting the need for our proposed task and approach.
Our designed \ModelNametrain{} model generates the most accurate predictions compared to baselines, demonstrating the effectiveness of our design.
Further qualitative examples and failure case analysis can be found in \secref{sec:add_qualitative} and \secref{sec:failure_cases} in the supplementary material.

We provide additional visualizations on the model's generalization capability in~\figref{fig:qual_generalize}.
The input images are randomly collected from online sources and are \textbf{not} part of the AGD20K dataset.
We observe that with the same input image, our model is sensitive to the provided affordance label and can accurately generate the corresponding affordance regions.
Notably, the model generalizes not only to novel affordance types on known object categories (\textbf{(a), (b)}), but also extends to truly novel combinations of novel object categories and novel affordance types (\textbf{(d)}).

\input{figs/qual_generalize}

\myparagraph{Discussion.}
Most affordance grounding and segmentation baselines struggle on our \benchmark{} benchmarks.
We identify two main challenges.
First, existing affordance grounding methods often assume fixed affordance types, making them either hard to adapt~\cite{chen2024worldafford, li2023locate, xu2024weakly, Learningluo} or prone to overfitting~\cite{li:ooal:2024}.
Second, affordance perception is not object recognition~\cite{gibson2014ecological}; VLMs and multimodal LLMs are primarily trained for object recognition and remain unreliable for \emph{direct} affordance generalization.
In contrast, our model better exploits pretrained VLMs through carefully designed feature representations and downstream architectures, providing a first step toward improving novel affordance generalization.

\begin{table}[t]
    \begin{minipage}[t]{0.39\linewidth}
        \input{tabs/ablation_trainfree}
    \end{minipage}
    \hfill
    \begin{minipage}[t]{0.58\linewidth}
        \input{tabs/ablation_trainable}
    \end{minipage}
\end{table}

\subsection{Ablation studies}
\label{sec:exp_ablation}

For \ModelNamefree{}, the experiments follow the setup in \secref{sec:exp_trainingfree}.
We report the results on \agdbenchmark{} val set in~\tabref{tab:ablation_trainfree}.
We start with only using the detection-and-segmentation backbone Grounded-SAM~\cite{ren2024grounded}.
We then evaluate the variant with semantic fusion only,~\ie, directly using the raw object mask $\mM^O_n$ for the semantic fusion instead of refined patches in $\gM^L_n$.
We observe that the complete \ModelNamefree{} with all designed modules achieves the best performance.

\input{tabs/ablation_branch}
For \ModelNametrain{}, the experiments follow the setup in~\secref{sec:exp_trainable}.
The results are shown in~\tabref{tab:ablation_trainable}.
We observe that incorporating the proposed trainable modules gradually improves the performance.
The complete \ModelNametrain{} with all submodules achieves the best overall performance.
We then show ablation studies for the multi-branch design in~\tabref{tab:ablation_branch}.
For the local-branch-only variant, when no object masks are detected, the model cannot produce a local prediction; in these cases, we return a uniform mask as a fallback.
Our complete model \ModelNametrain{}, which combines both branches, achieves the best overall performance.
For additional ablation studies, please refer to Appendix~\secref{sec:ablation}.

%% file: tabs/main_training_free.tex
\centering
\small
\caption{
\textbf{Quantitative comparison for the training-free setting.} 
We report the IoU@0.4 ($\uparrow$) on \agdbenchmark{} and \umdbenchmark{}.
`AA.' denotes our \ModelNamefree{}.
}
    \begin{tabular}{l cc}
    \specialrule{.15em}{.05em}{.05em}
    Method & AGD. & UMD. \\
    \cline{1-3}
    SCLIP~\cite{wang2024sclip} & 4.2 & 2.0 \\
    ClearCLIP~\cite{lan2024clearclip} &  4.4 & 0.5 \\
    ProxyCLIP~\cite{lan2024proxyclip} &  6.4 & 4.0\\
    GEM~\cite{bousselham2024grounding} &  8.2 & 0.9 \\
     \colortrainfree AA. & \textbf{21.3} & \bf 27.9
    \\
    \specialrule{.15em}{.05em}{.05em}
    \end{tabular}
\label{tab:main_training_free}

%% file: tabs/main_trained.tex
\centering
\caption{
\textbf{Quantitative comparison for the one-shot training setting.} 
We report KLD, SIM, NSS, and IoU@0.4 on \agdbenchmark{}; and IoU@0.4 on \umdbenchmark{}.
`AA+.' denotes our \ModelNametrain{}.
}
\resizebox{1\linewidth}{!}{
    \begin{tabular}{lcccccc}
    \specialrule{.15em}{.05em}{.05em}
    & \multicolumn{4}{c}{AGD.} & UMD. \\
    \cline{2-6}
    \multirow{-2}{*}{Method} & KLD($\downarrow$) & SIM($\uparrow$) & NSS($\uparrow$) & IoU($\uparrow$) & IoU($\uparrow$) \\ 
    \hline
    SCLIP~\cite{wang2024sclip} & 2.057 & 0.215 & 0.189 & 9.3 & 5.5 \\
    ClearCLIP~\cite{lan2024clearclip} & 1.722 & 0.269 & 0.803 & 13.6 & 8.8 \\
    ProxyCLIP~\cite{lan2024proxyclip} &  1.925 & 0.245 &  0.459 & 13.7 & 6.9 \\
    GEM~\cite{bousselham2024grounding} & 1.860 & 0.247 & 0.516 & 11.6 & 26.9 \\        
    OOAL~\cite{li:ooal:2024} & 1.617 & 0.323 & 0.803 & 10.6 & 18.3 \\
    \colortrainable {AA+.} & \textbf{1.374} & \textbf{0.368} & \textbf{1.053} & \textbf{22.9} & \bf 30.3 \\
    \specialrule{.15em}{.05em}{.05em}
    \end{tabular}
}
\label{tab:main_trained}

%% file: tabs/add_datasplit_results.tex
\begin{wraptable}[9]{r}{0.55\linewidth}
\centering
\vspace{-0.45cm}
\caption{
\textbf{Additional quantitative comparison on different data splits.}
‘A1/2’ denotes \emph{Additional split 1/2}.
`AA+.' represents our \ModelNametrain{}.
}
\resizebox{1\linewidth}{!}{
    \begin{tabular}{l l c c c c}
    \specialrule{.15em}{.05em}{.05em}
    Split & Method & KLD ($\downarrow$) & SIM ($\uparrow$) & NSS ($\uparrow$) & IoU@0.4 ($\uparrow$) \\
    \hline
    \multirow{2}{*}{A1} 
      & OOAL~\cite{li:ooal:2024} 
      & 1.401 & 0.394 & 1.042 & 20.4 \\
      & \cellcolor{trainable}{AA+.}
      & \cellcolor{trainable}\textbf{1.231} 
      & \cellcolor{trainable}\textbf{0.404} 
      & \cellcolor{trainable}\textbf{1.221} 
      & \cellcolor{trainable}\textbf{29.5} \\
    \hline
    \multirow{2}{*}{A2} 
      & OOAL~\cite{li:ooal:2024} 
      & 1.230 & 0.424 & 1.097 & 14.0 \\
      & \cellcolor{trainable}{AA+.}
      & \cellcolor{trainable}\textbf{1.082} 
      & \cellcolor{trainable}\textbf{0.453} 
      & \cellcolor{trainable}\textbf{1.275} 
      & \cellcolor{trainable}\textbf{29.3} \\
    \specialrule{.15em}{.05em}{.05em}
    \end{tabular}
}
\label{tab:add_datasplit_results}
\vspace{-0.5cm}
\end{wraptable}

%% file: figs/qualitative.tex
\begin{figure}[t]
\centering
\includegraphics[width=1\linewidth]{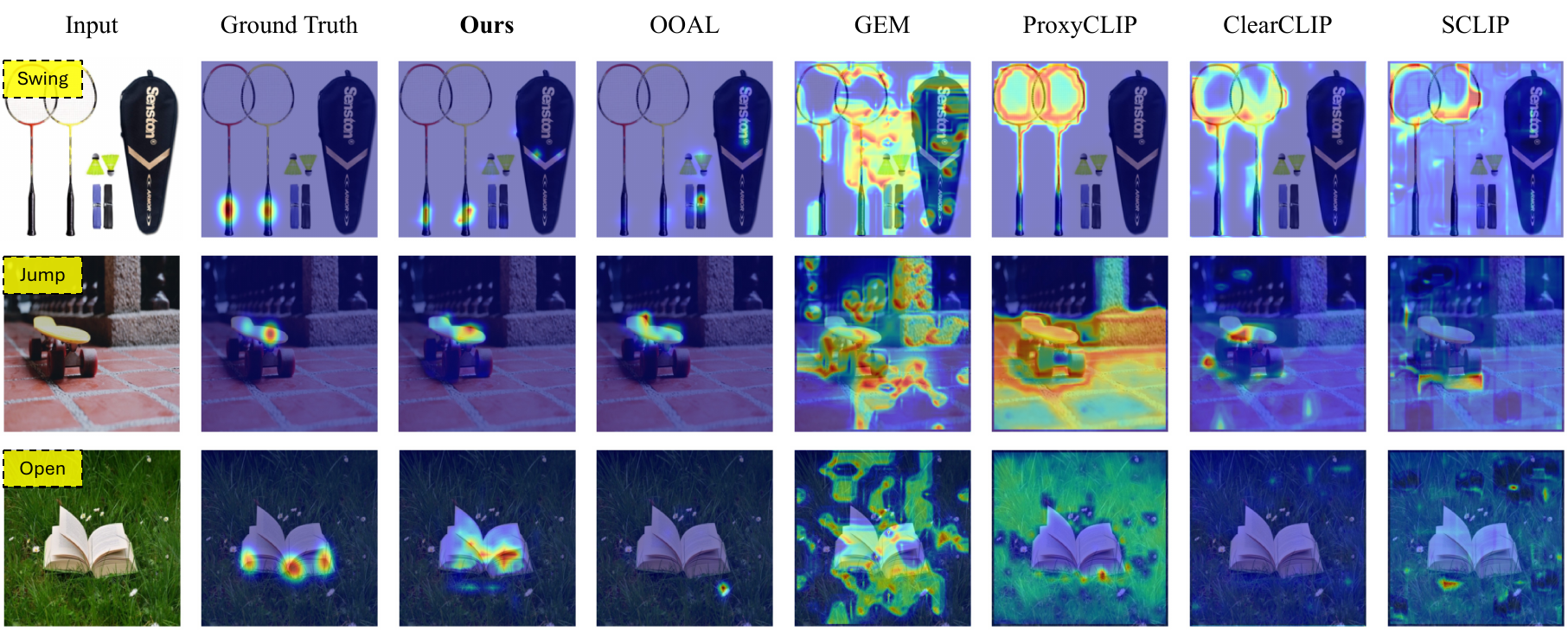}
\vspace{-0.45cm}
\caption{\textbf{Qualitative results on \agdbenchmark{} test set.}
We compare our \ModelNametrain{} against other affordance grounding and fine-tuned open-vocabulary semantic segmentation baselines.
Our method consistently localizes affordance-relevant subregions for novel affordance types.
}
\vspace{-0.4cm}
\label{fig:qualitative}
\end{figure}

%% file: figs/qual_generalize.tex
\begin{figure}[t]
    \centering
    \includegraphics[width=1.03\linewidth]{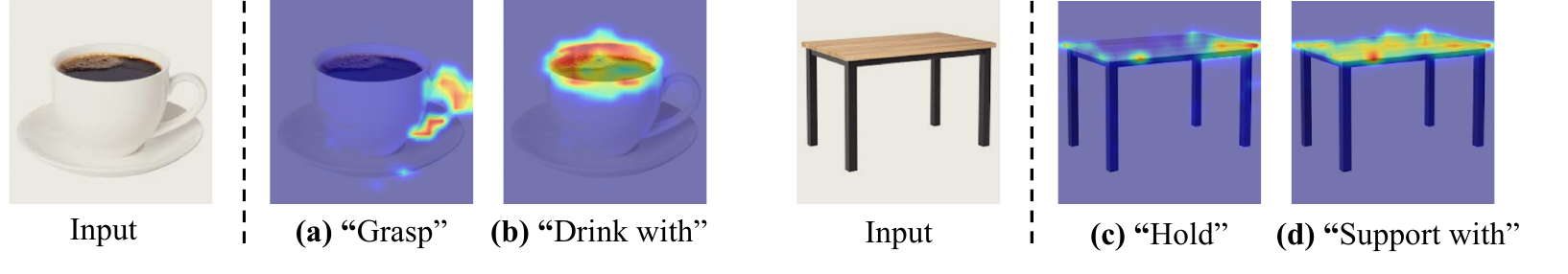}
    \vspace{-0.5cm}
    \caption{\textbf{Additional qualitative results on generalization.}
    We show the predictions on {\bf (a)(b)} base object categories but \textit{novel} affordances; {\bf (c)} \textit{novel} object categories but base affordances; and {\bf (d)} \textit{novel} object categories and \textit{novel} affordances.}
    \label{fig:qual_generalize}
\end{figure}

%% file: tabs/ablation_trainfree.tex
\centering
\caption{
\textbf{Ablation studies for the proposed modules in \ModelNamefree{}}. 
`D\&S.' stands for the detection-and-segmentation backbone. 
`SF.' stands for the semantic fusion module. 
`PR.' stands for the patch refinement module.
}
    \begin{tabular}{cccc}
    \specialrule{.15em}{.05em}{.05em}
     D\&S. & SF. & PR. & IoU@0.4($\uparrow$) \\
    \hline
     \ccmark & \cxmark & \cxmark & 17.9 \\
     \ccmark & \ccmark & \cxmark & 18.5 \\
    \colortrainfree \ccmark & \ccmark & \ccmark & \textbf{20.5} \\
    \specialrule{.15em}{.05em}{.05em}
    \end{tabular}
\label{tab:ablation_trainfree}

%% file: tabs/ablation_trainable.tex
\centering
\caption{\textbf{Ablation studies for the proposed modules in the trainable fusion module}.
`GB.' stands for global branch decoder. 
`BW.' stands for branch weight predictor. 
`LW.' stands for local weight predictor.
The `\cxmark' sign indicates %
that the corresponding module is replaced by its training-free \ModelNamefree{} counterpart.
}
\resizebox{1\linewidth}{!}{
    \begin{tabular}{ccccccc}
    \specialrule{.15em}{.05em}{.05em}
     GB. & BW. & LW. & KLD($\downarrow$) & SIM($\uparrow$) & NSS($\uparrow$) & IoU@0.4($\uparrow$) \\
    \hline
    \colortrainfree \cxmark & \cxmark & \cxmark & 6.186 & 0.303 & 0.430 & 20.5 \\
     \ccmark & \cxmark & \cxmark & 5.037 & 0.386 & 1.100 & 25.9 \\
     \ccmark & \ccmark & \cxmark & 1.196 & 0.439 & 1.274 & 32.5 \\
    \colortrainable \ccmark & \ccmark & \ccmark & \textbf{1.150} & \textbf{0.460} & \textbf{1.316} & \textbf{34.2} \\
    \specialrule{.15em}{.05em}{.05em}
    \end{tabular}
}
\label{tab:ablation_trainable}

%% file: tabs/ablation_branch.tex
\begin{wraptable}[7]{r}{0.57\textwidth}
\centering
\vspace{-0.45cm}
\caption{\textbf{Ablation studies for the multi-branch design}. `Glo.'/`Loc.' stands for global/local branch.
}
\resizebox{1\linewidth}{!}{
    \begin{tabular}{cccccc}
    \specialrule{.15em}{.05em}{.05em}
     Glo. & Loc. & KLD($\downarrow$) & SIM($\uparrow$) & NSS($\uparrow$) & IoU@0.4($\uparrow$) \\
    \hline
     \cxmark & \ccmark & 10.435 & 0.268 & 0.629 & 18.5 \\
     \ccmark & \cxmark & 1.170 & 0.449 & 1.244 & 32.9 \\
    \colortrainable \ccmark & \ccmark & \textbf{1.150} & \textbf{0.460} & \textbf{1.316} & \textbf{34.2} \\
    \specialrule{.15em}{.05em}{.05em}
    \end{tabular}
}
\label{tab:ablation_branch}
\end{wraptable}

%% file: sec/conclusion.tex
\section{Conclusion}
\label{sec:conclusion}

This work introduces zero-shot 2D grounding of novel affordance types, studying a critical gap in the ability of existing models to generalize beyond fixed affordance label sets.
To facilitate research, we establish the \benchmark{} benchmarks, providing a systematic evaluation protocol for this challenging setting.
We propose \ModelNamefree{}, a training-free pipeline that leverages segmentation cues to localize affordance regions.
We further introduce \ModelNametrain{}, an extension that incorporates a trainable fusion module to enhance performance.
Experiments show that the \benchmark{} benchmarks pose significant challenges for current methods, and our approach achieves up to a 12.3\% absolute IoU@0.4 improvement over prior work.
We hope these benchmarks and findings encourage a more systematic study of generalization in 2D affordance grounding.

%% file: supp_src/suppl.tex
\clearpage
\setcounter{page}{1}

\setcounter{section}{0}
\renewcommand{\theHsection}{A\arabic{section}}
\renewcommand{\thesection}{A\arabic{section}}
\renewcommand{\thetable}{A\arabic{table}}
\setcounter{table}{0}
\setcounter{figure}{0}
\renewcommand{\thetable}{A\arabic{table}}
\renewcommand\thefigure{A\arabic{figure}}
\renewcommand{\theHtable}{A.Tab.\arabic{table}}%
\renewcommand{\theHfigure}{A.Abb.\arabic{figure}}%
\renewcommand\theequation{A\arabic{equation}}
\renewcommand{\theHequation}{A.Abb.\arabic{equation}}%

\section*{Appendix}

\noindent The appendix is organized as follows:
\begin{itemize}
\item In~\secref{sec:datasplit}, we provide details on the datasets and splits for the proposed \benchmark{} benchmarks.
\item In~\secref{sec:implementation_details}, we provide implementation details on our proposed models.
\item In~\secref{sec:baseline_models}, we provide implementation details on the baseline models.
\item In~\secref{sec:eval_metrics}, we provide details on the evaluation metrics.
\item In~\secref{sec:add_existing}, we provide additional quantitative results on existing affordance grounding benchmarks.
\item In~\secref{sec:ablation}, we provide additional ablation studies on our proposed models.
\item In~\secref{sec:hyperparameter}, we provide hyperparameter search details on our proposed models.
\item In~\secref{sec:computation}, we provide the computational cost of our proposed model.
\item In~\secref{sec:add_qualitative}, we provide additional qualitative results on the \agdbenchmark{} benchmark.
\item In \secref{sec:limitation}, we discuss the limitations of our proposed methods.
\item In \secref{sec:failure_cases}, we discuss some failure cases of our proposed methods.
\end{itemize}

\section{Dataset and Data Splits}
\label{sec:datasplit}
We now provide more details on the datasets and data splits for our \agdbenchmark{} and \umdbenchmark{} benchmarks. Code and data will be made publicly available. 

{\bf \noindent Datasets.} The AGD20K dataset is a large-scale affordance grounding dataset for weakly-supervised learning. 
It contains 20,061 exocentric images and 3,755 egocentric views, spanning 36 affordance types and 50 object classes.
The UMD part affordance dataset contains 28,843 RGB-D images, spanning 7 affordances and 17 object categories.
The ground truth affordance region heatmaps for the AGD20K dataset are generated by applying a Gaussian blur over the annotated keypoints in the corresponding affordance region, while the UMD dataset is annotated with dense binary maps at the pixel level.

{\bf \noindent \agdbenchmark{}.}
We report the affordance type splits for \agdbenchmark{} in \tabref{tab:supp_datasplits}. 
Following the existing benchmark setting~\cite{Learningluo} for novel object categories, we only use the affordance types shared across multiple objects.
For the training data, we adopt the one-shot training setup introduced by OOAL~\cite{li:ooal:2024}, where 50 randomly selected egocentric images per object category from AGD20K are annotated.
We decompose each image annotation into multiple image-affordance pairs and select only the pairs where the affordance types belong to the divided training split.
Note that this uses fewer training samples than OOAL.
For the validation and test data, we use the entire AGD20K test set in the `Seen' setting and further split it following the divided validation and test affordance types.
In total, we have 63 image-affordance pairs for training, 208 pairs for validation, and 345 pairs for testing.

{\bf \noindent \umdbenchmark{}.}
We report the affordance type splits for \umdbenchmark{} in~\tabref{tab:supp_datasplits}. 
Similar to \agdbenchmark{}, for each image annotation, we decompose it into multiple image-affordance pairs and only retain the pairs where the affordance types belong to the divided training split.
For the validation and test data, we randomly select up to 20 samples per object in a scene to reduce redundancy, as the images captured within the same scene tend to be highly similar.
In total, we have 156 image-affordance pairs for training, 482 pairs for validation, and 760 pairs for testing.

{\bf \noindent One-shot setting.}
For both benchmarks, we follow the one-shot training setup from prior work~\cite{li:ooal:2024}, where only a single annotation is used per object during training.
For the \agdbenchmark{} benchmark, this corresponds to one annotated image per object–affordance combination, while in \umdbenchmark{}, only one annotation per object is used.

Note that in the prior one-shot setting~\cite{li:ooal:2024}, a single annotation may contain multiple affordance maps corresponding to different affordance types.
When using image–affordance pairs as input, each object can therefore appear multiple times, each associated with a distinct affordance type.
Similarly, in our setting, each affordance type may appear multiple times within different image-affordance pairs.
To be consistent with the one-shot definition of prior work~\cite{li:ooal:2024}, we adopt the same protocol in our benchmarks.
Nevertheless, our data split in~\tabref{tab:supp_datasplits} could also be extended to other training settings, such as the weakly supervised setting.

\input{tabs/supp_datasplits}
\input{tabs/supp_existing}

\section{Implementation Details of Our Models}
\label{sec:implementation_details}
In this section, we provide the implementation details of the proposed \ModelNamefree{} and \ModelNametrain{}.

\myparagraph{Local patch construction.}
Given an object mask $\mM_n^O$ of size $H \times W$, we construct local patches $\mM_n^L$ using both evenly-focused and border-focused crops.
For evenly-focused patches $\mM_n^E$, we uniformly split the mask into a $g \times g$ grid, with each cell covering $\frac{1}{g}H \times \frac{1}{g}W$.
This yields $e = g \times g$ evenly-focused patches in total.
For border-focused patches $\mM_n^B$, we use a border ratio $b$ and split the image coordinates at $\{0,bW,(1-b)W,W\}$ horizontally and $\{0,bH,(1-b)H,H\}$ vertically.
This yields a $3 \times 3$ grid whose outer cells emphasize object boundary regions and whose center cell covers the remaining interior.
Each local patch is obtained by cropping the corresponding grid cell from the object-masked image.
We combine the full object crop, $\mM_n^E$, and $\mM_n^B$ as the final local patch set $\mM_n^L$.

{\bf \noindent \ModelNamefree{}.}
For open-vocabulary detection and segmentation, we adopt Grounded-SAM~\cite{ren2024grounded}, with GroundingDINO~\cite{liu2024grounding} and SAM2~\cite{ravi2024sam} as the backbone. 
For feature extraction, we use a frozen pre-trained CLIP~\cite{radford2021learning} with ViT-B/32. 
For the patch refinement of local patches generation, we set the number of evenly-focused patches $e$ to 9 and the border ratio $b$ to 0.2.
We set the scaling factor $d$ to 512, which is the dimension of the CLIP features.
\new{We set the default branch weight $\alpha$ in~\equref{eqn:branch_weight} to 0, and assign it to 1 when no objects are detected by Grounded-SAM.}

{\noindent \ModelNametrain{}.}
For the pre-trained backbones, we replace CLIP with DINO to extract more comprehensive image features $\vf^I$, while keeping all other settings consistent with \ModelNamefree{}.
For the transformer layers in the trainable fusion module, we follow the same settings in terms of dimension size, layer number, and head size as used for the affordance grounding baseline~\cite{li:ooal:2024}. 
For \agdbenchmark{} and \umdbenchmark{}, we train the model using the Adam optimizer with a learning rate of 0.0001 and 0.00001 for 50 epochs, respectively.
We select the best model based on the corresponding metrics on the validation set.
For \agdbenchmark{}, we choose the model based on KLD.
For \umdbenchmark{}, we choose the model based on IoU@0.4.
We use binary cross-entropy (BCE) as the training loss.
Given the ground truth map $\mY$ and the predicted map $\hat{\mY}$, the BCE loss is calculated as:
\bea
        \mathcal{L}_{\text{BCE}}=%
        -\sum_i(\mY_i \log \hat{\mY_i} + (1- \mY_i) \log(1-\hat{\mY_i})).
\eea

\section{Implementation Details for Baselines}
\label{sec:baseline_models}
We provide additional implementation details of baseline models. For OOAL~\cite{li:ooal:2024}, we modify the model so that it generates the corresponding text embeddings for each new affordance type input.
For \agdbenchmark{}, we follow the open-source training script to train the model using SGD optimizer with a learning rate of 0.01 for 20k iterations.
For \umdbenchmark{}, we train with a learning rate of 0.001.

For GEM~\cite{bousselham2024grounding}, we add a prediction head and set the self-attention layers and residual blocks to be trainable when evaluating in the one-shot training setting.
The additional prediction head consists of 3 convolutional layers and 2 batch norm layers. The input and output channels for each convolutional layer are set to (1, 32), (32, 32), (32, 1), respectively. The kernel size is set to 3, and the padding size is set to 1.
For \agdbenchmark{}, we train the model using the Adam optimizer with a learning rate of 0.01 for 50 epochs.
For \umdbenchmark{}, we train the model with a learning rate of 0.001.

For SCLIP~\cite{wang2024sclip}, ClearCLIP~\cite{lan2024clearclip}, and ProxyCLIP~\cite{lan2024proxyclip}, we apply the same prediction head used in GEM.
For \umdbenchmark{}, we add a `background' category as a placeholder, since there is only one affordance type for the test set.
For \agdbenchmark{}, we train the models using the Adam optimizer with a learning rate of 0.001 for 50 epochs.
For \umdbenchmark{}, we train the models with a learning rate of 0.01.
For open-vocabulary segmentation models, we provide the affordance type as the text prompt.

\new{For \texttt{GPT-5.2}~\cite{openai2026gpt52}, we query the model with the input image and the prompt: \textit{``I want to infer the corresponding affordance region for the input image. Please output the affordance region in bounding boxes. The affordance label is \{AFF\}. Please directly output text without generating new images. The output should be in the format [ymin, xmin, ymax, xmax].''},
where we provide the affordance label in \textit{\{AFF\}}.}

\section{Evaluation Metrics}
\label{sec:eval_metrics}
We provide details on the evaluation metrics.
We evaluate Kullback-Leibler Divergence (KLD), Similarity (SIM), Normalized Scanpath Saliency (NSS), and Intersection-over-Union at threshold K (IoU@K).

{\bf \noindent KLD.} 
Given the ground truth map $\mY^{\text{gt}}$ and the predicted map $\mY^{\text{pred}}$, we first compute the normalized maps:
\bea
\hat{\mY}^{\text{gt}}_i = \frac{\mY^{\text{gt}}_i}{\sum_{i'}{\mY_{i'}^{\text{gt}}}} ~\text{ and }~
\hat{\mY}^{\text{pred}}_i = \frac{\mY^{\text{pred}}_i}{\sum_{i'}{\mY_{i'}^{\text{pred}}}}.
\eea
The KLD of the normalized maps is evaluated as:
\bea
\text{KLD}(\hat{\mY}^{\text{gt}} || \hat{\mY}^{\text{pred}}) = \sum_{i}{\hat{\mY}_i^{\text{gt}} \cdot \operatorname{log}\frac{\hat{\mY}_i^{\text{gt}}}{\hat{\mY}_i^{\text{pred}}}}.
\eea

{\bf \noindent SIM.}
The SIM of the normalized maps is evaluated as:
\bea
\text{SIM}(\hat{\mY}^{\text{gt}}, \hat{\mY}^{\text{pred}}) = \sum_{i}{\operatorname{min}(\hat{\mY}_i^{\text{gt}}, \hat{\mY}_i^{\text{pred}})}.
\eea

{\bf \noindent NSS.}
Given the ground truth map $\mY^{\text{gt}}$ and the predicted map $\mY^{\text{pred}}$, we first compute the following maps:
\bea
\overline{\mY}_i^{\text{gt}}=\mathbbm{1}[\mY_i^{\text{gt}}>0.1] ~\text{ and }~ \overline{\mY}_i^{\text{pred}}=\frac{\mY_i^{\text{pred}}-\mu(\mY^{\text{pred}})}{\sigma(\mY^{\text{pred}})},
\eea
where $\mathbbm{1}[\cdot]$ is an indicator function, $\mu$ denotes the mean of the elements in the map, and $\sigma$ denotes the standard deviation.

The NSS of the normalized maps is evaluated as:
\bea
\text{NSS}(\overline{\mY}^{\text{gt}}, \overline{\mY}^{\text{pred}}) = \frac{1}{\sum_{i'}\overline{\mY}_{i'}^{\text{gt}}}\sum_{i}{\overline{\mY}_i^{\text{gt}} \cdot \overline{\mY}_i^{\text{pred}}}.
\eea

{\bf \noindent IoU@K.}
Given the ground truth map $\mY^{\text{gt}}$ and the predicted map $\mY^{\text{pred}}$, we first generate the binary masks $\mY^{\text{gt}'}$ and $\mY^{\text{pred}'}$ based on the threshold K:
\bea
\mY_i^{\text{gt}'}=\mathbbm{1}[\mY_i^{\text{gt}}>K], \quad 
\mY_i^{\text{pred}'}=\mathbbm{1}[\mY_i^{\text{pred}}>K].
\eea
IoU@K is evaluated as:
\bea
\text{IoU@K}(\mY^{\text{gt}}, \mY^{\text{pred}}) = \frac{|\mY^{\text{gt}'} \cap \mY^{\text{pred}'}|}{|\mY^{\text{gt}'} \cup \mY^{\text{pred}'}|},
\eea
where $|\cdot|$ indicates the area.

\section{Additional Quantitative Results on Existing Affordance Grounding Benchmarks}
\label{sec:add_existing}
In this section, we provide additional quantitative results on existing affordance grounding benchmarks, \ie, the ``Seen'' and ``Unseen'' settings in AGD20K benchmark~\cite{Learningluo}.
The results are shown in~\tabref{tab:supp_existing}.
The experiment follows the one-shot setting in~\citet{li:ooal:2024}.

We observe that our proposed \ModelNametrain{} achieves competitive performance on the AGD20K benchmark, where the training and testing affordance types are from a pre-defined closed set.
With either CLIP or DINO as the backbone, the prior state-of-the-art affordance grounding model OOAL~\cite{li:ooal:2024} still achieves better performance.
We attribute this to its learnable prompt design, which is effective under the assumption of fixed affordance types.
However, as shown in~\tabref{tab:main_trained}, this assumption limits performance when the affordance types are open-set. Additionally, we find that fine-tuning the training-free open-vocabulary segmentation models is insufficient for the affordance grounding task, consistent with findings in~\tabref{tab:main_trained} and~\citet{li:ooal:2024}.

\new{Empirically, OOAL achieves better KLD on novel object categories (1.070) than on novel affordance types (1.617 in~\tabref{tab:main_trained}), although the comparison is not strictly fair due to different data splits. This observation further highlights the difficulty of generalization across affordance types. We attribute this gap to the fact that object categories often exhibit strong visual cues (\eg, shape or texture) that facilitate recognition, whereas affordance types are more abstract and less directly tied to appearance.}

\input{tabs/supp_ablation_mr_patch}
\input{tabs/supp_ablation_backbone}
\input{tabs/supp_ablation_text}
\input{tabs/supp_ablation_pl}

\section{Additional Ablation Studies}
\label{sec:ablation}
In this section, we provide additional ablation studies for our proposed models.
The results are reported on the \agdbenchmark{} validation set.

{\bf \noindent Detection module.}
We conduct ablation studies for the detection module of the detection-and-segmentation backbone. 
We compare our \ModelNamefree{} with the model that removes the detection backbone GroundingDINO, \ie, directly applying SAM2 to the input image $\mI$ to obtain the mask set $\gM^O$.
The experimental setting follows the training-free setting on the \agdbenchmark{} benchmark in~\secref{sec:exp_trainingfree}.
We find that the IoU@0.4 metric of the segmentation-only backbone is only 7.5, compared to 20.5 for our designed \ModelNamefree{} with the detection-and-segmentation backbone.
This result suggests the importance of locating the object first in the affordance grounding task and validates our design.

{\bf \noindent Detection-and-segmentation backbone.}
We conduct ablation studies for different detection-and-segmentation backbones. 
We compare our \ModelNamefree{} with the variant that uses UNINEXT-R~\citep{yan2023universal} as the detection-and-segmentation backbone.
The experimental setting follows the training-free setting on the \agdbenchmark{} benchmark in~\secref{sec:exp_trainingfree}.
Results show that the UNINEXT-R backbone achieves an IoU@0.4 of only 13.7, whereas Grounded-SAM~\cite{ren2024grounded} yields 20.5. These findings indicate that employing a stronger detection-and-segmentation backbone leads to better affordance grounding performance.

We further analyze the reliability of the detection-and-segmentation backbone.
A detection is considered successful if the predicted object mask overlaps with the ground-truth affordance annotation, \ie, the detected object contains the target affordance region.
Grounded-SAM returns object detections in 281 out of 345 test samples, among which 277 overlap with the ground-truth affordance regions.
The object-level precision is 98.58\% and recall is 80.29\%.

{\bf \noindent Patch Refinement Module.}
We ablate on the designed patches in the patch refinement module, as shown in \tabref{tab:supp_ablation_mr_patch}.
The experimental setting follows the one-shot setting on the \agdbenchmark{} benchmark in \secref{sec:exp_trainable}.
We observe that similar to the \ModelNamefree{} results in \tabref{tab:ablation_trainfree}, the patch refinement module improves the performance over the variant without it.
Each type of the designed patch contributes to the performance gain, and using all patches yields the best result.

{\bf \noindent Image encoder.} 
We provide the ablation studies for the image backbone, as shown in~\tabref{tab:supp_ablation_backbone}.
Our proposed \ModelNametrain{} consistently outperforms the SOTA affordance grounding method OOAL~\cite{li:ooal:2024} regardless of the image backbone.
Comparing with the result in~\tabref{tab:main_trained}, we observe that even with CLIP as the backbone, our proposed \ModelNametrain{} consistently outperforms all baselines across all metrics on both benchmarks.
Replacing CLIP with DINO as the backbone further boosts the performance, which aligns with the observation in~\citet{li:ooal:2024}.

{\bf \noindent Text encoder.}
To validate whether a more advanced text encoder would lead to a straightforward performance improvement, we provide additional experimental results with ViT‑g/14 as the text encoder in~\tabref{tab:supp_ablation_text}.
We observe that directly applying a stronger text encoder did not lead to a clear performance improvement. This suggests that our task is less constrained by text representations and more reliant on robust visual representations and effective grounding mechanisms.

{\bf \noindent Prompt learner.} 
We provide the ablation studies for the text prompt learner proposed by~\citet{li:ooal:2024}, as shown in~\tabref{tab:supp_ablation_pl}.
The experimental setting follows the one-shot setting on the \agdbenchmark{} benchmark in \secref{sec:exp_trainable}.
As our setting does not assume a predefined affordance set, we use an empty string as the placeholder.
Comparing Rows 1 \& 2 and Rows 3 \& 4, we confirm that removing the prompt learner improves the performance for both methods, suggesting that the prompt learner hurts generalization to novel affordances. 
Comparing Rows 1 \& 3 and Rows 2 \& 4, we verify that our carefully designed visual modules bring substantial improvement, independent of prompt learning.

\section{Hyperparameter Search}
\label{sec:hyperparameter}
We provide details on the hyperparameter search conducted for our proposed models.

{\bf \noindent Patch refinement module.}
We conduct a hyperparameter search for the proposed patch refinement module.
Specifically, we examine the effect of two parameters: the number of evenly-focused patches $e$ and the border ratio $b$ in the patch refinement module of \ModelNamefree{}.
The experimental setting follows the training-free setting on the \agdbenchmark{} benchmark in~\secref{sec:exp_trainingfree}.
The results are shown in~\tabref{tab:supp_hyper_postprocess}.
We observe that the model performance is robust across different combinations of parameters.
This validates our idea of identifying local regions to improve affordance grounding.
Based on these results, we set the number of evenly-focused patches $e$ to 9 and the border ratio $b$ to 0.2 for both \ModelNamefree{} and \ModelNametrain{}.

{\bf \noindent Branch initialization.}
We find that branch initialization has a significant impact on the training performance, as shown in \tabref{tab:supp_hyper_init}.
The experimental setting follows the one-shot setting on the \agdbenchmark{} benchmark in \secref{sec:exp_trainable}.
When $\alpha$ is initialized to 1, the model mainly relies on the global branch prediction, leading to slightly worse performance.
When initializing $\alpha$ to 0, the model depends mainly on the refined training-free local segmentation cues, resulting in a substantial drop in overall performance.
Initializing the model to equally attend to both branches yields the best performance, allowing balanced contributions from global and local branches.

{\bf \noindent Learning rate.}
We conduct a hyperparameter search over the learning rate, shown in~\tabref{tab:supp_hyper_lr}.
The experimental setting follows the one-shot setting on the \agdbenchmark{} benchmark in \secref{sec:exp_trainable}.
We observe that a learning rate of 0.0001 yields the best performance across most metrics.
We choose the learning rate based on the best performance over the KLD metric.

\input{tabs/supp_hyper_postprocess}
\input{tabs/supp_hyper_init}
\input{tabs/supp_hyper_lr}

\input{tabs/supp_computation}

\section{Computational Cost}
\label{sec:computation}
We now analyze the computational cost of our designed \ModelNametrain{}.
We report the number of trainable parameters and the inference time of \ModelNametrain{} and the baseline model OOAL in~\tabref{tab:supp_computation}. 
All experiments are conducted on the \agdbenchmark{} test set on a single NVIDIA A100 GPU.

Our method's overall trainable parameters are comparable to the baseline. For total inference time, our \ModelNametrain{} takes longer than the baseline model OOAL~\cite{li:ooal:2024}.
We also report the detailed inference time for the local patches generation procedure (pre-processing), which includes the detection-and-segmentation backbone and the patch refinement process. The main contributor to the longer total inference time comes from the local patches generation.

\input{figs/gpt5}
\input{figs/supp_qualitative}

\section{Additional Qualitative Results}
\label{sec:add_qualitative}
We provide additional qualitative results on the \agdbenchmark{} benchmark, as shown in~\figref{fig:supp_qualitative}.
Both OOAL~\citep{li:ooal:2024} and our proposed \ModelNametrain{} are trained with a CLIP backbone.
Consistent with our observations in~\secref{sec:exp_trainable}, we find that most existing baselines struggle to localize regions that correspond to the unseen affordance types.
Our designed \ModelNametrain{} demonstrates better performance.

\section{Limitations \& Future Directions}
\label{sec:limitation}
For the benchmarks, we currently provide a single primary data split shown in~\secref{sec:datasplit}, with two additional splits evaluated in~\secref{sec:exp_trainable}.
A promising future direction is to systematically analyze how different affordance type partitions influence the generalization performance.
Moreover, similar to prior work~\cite{li:ooal:2024}, the limited availability of dense annotations (\secref{sec:datasplit}) makes it challenging to comprehensively assess robustness across varying training data.

For the designed models, we incorporate a pre-trained vision-language model for detection and segmentation, which means the performance depends on the capabilities of the backbone.
Additionally, since we need to generate the local patches, our overall inference time is longer compared to existing affordance grounding models.
For more details on the computational cost, please refer to~\secref{sec:computation}.

\input{figs/supp_failure_cases}

\section{Failure cases}
\label{sec:failure_cases}
We visualize several failure cases of \ModelNametrain{} on the \agdbenchmark{} test set in~\figref{fig:supp_failure_cases}.
We observe that the model may struggle in scenarios where the target affordance corresponds to multiple disjoint regions on the same object.
Since these regions can vary in size and spatial configuration, the current patch refinement module may fail to capture fine-grained details that extend beyond individual grid boundaries.
A future direction is to explore more flexible patch refinement mechanisms to model complex, spatially distributed affordance patterns.

\input{tabs/supp_add_datasplits}

\input{tabs/rel_work}

%% file: tabs/supp_datasplits.tex
\begin{table*}[t]
\centering
\caption{\textbf{Data splits for the proposed \agdbenchmark{} and \umdbenchmark{}
benchmarks.}}
\resizebox{1\textwidth}{!}{
    \begin{tabular}{lccc}
    \specialrule{.15em}{.05em}{.05em}
        Benchmarks & Train & Val & Test \\
        \hline
        & carry, eat, hold, kick, pick\_up & text\_on, stick
        & push, drink\_with, brush\_with, lift\\
        & take\_photo, throw, wash, peel, pour & cut, hit
        & boxing, catch, open, jump \\
        \multirow{-3}{*}{\agdbenchmark{}} & sip, talk\_on, drag, pack, ride, sit\_on & lie\_on
        & cut\_with, stir, swing \\
        \hline
        \umdbenchmark{} & grasp, cut, scoop, wrap\_grasp & pound, support & contain \\
    \specialrule{.15em}{.05em}{.05em}
    \end{tabular}
}
\label{tab:supp_datasplits}
\end{table*}

%% file: tabs/supp_existing.tex
\begin{table*}[t]
\centering
\caption{\textbf{Quantitative comparison for both `Seen' and `Unseen' settings on the AGD20K benchmark test set.}
Note that existing ``Unseen'' setting in AGD20K benchmark refers to unseen \textit{object categories}.
For OOAL~\cite{li:ooal:2024} and our proposed \ModelNametrain{}, we also report the results using CLIP as backbone for fair comparison.
}
    \begin{tabular}{lccccccc}
    \specialrule{.15em}{.05em}{.05em}
    & & \multicolumn{3}{c}{Seen} & \multicolumn{3}{c}{Unseen} \\
    \cline{3-8}
    \multirow{-2}{*}{Method} & \multirow{-2}{*}{Backbone} & KLD($\downarrow$) & SIM($\uparrow$) & NSS($\uparrow$) & KLD($\downarrow$) & SIM($\uparrow$) & NSS($\uparrow$) \\ 
    \hline
    SCLIP~\cite{wang2024sclip} & CLIP & 1.874 & 0.242 & 0.426 & 2.036 & 0.210 & 0.438 \\
    ClearCLIP~\cite{lan2024clearclip} & CLIP & 1.639 & 0.299 & 0.786 & 1.770 & 0.269 & 0.841 \\
    ProxyCLIP~\cite{lan2024proxyclip} & CLIP &1.717 & 0.293 & 0.717 & 1.887 & 0.256 & 0.688 \\
    GEM~\cite{bousselham2024grounding} & CLIP & 1.683 & 0.290 & 0.699 & 1.919 & 0.243 & 0.604 \\
    \hline
    & CLIP & 1.131 & 0.450 & 1.318 & 1.505 & 0.359 & 1.047 \\
    \multirow{-2}{*}{OOAL~\cite{li:ooal:2024}} & DINO & \textbf{0.740} & \textbf{0.577} & \textbf{1.745} & \textbf{1.070} & \textbf{0.461} & \textbf{1.503} \\
    \colortrainable & CLIP & 1.247 & 0.392 & 1.226 & 1.581 & 0.313 & 0.999 \\
    \colortrainable \multirow{-2}{*}{\ModelNametrain{}} & DINO & 0.885 & 0.516 & 1.575 & 1.272 & 0.411 & 1.284 \\
    \specialrule{.15em}{.05em}{.05em}
    \end{tabular}
\label{tab:supp_existing}
\end{table*}

%% file: tabs/supp_ablation_mr_patch.tex
\begin{table}[t]
\centering
\caption{\textbf{Ablation studies for the proposed patch types in the patch refinement module.} 
`OM.' stands for the original object mask. 
`EP.' stands for evenly-focused patches. 
`BP.' stands for border-focused patches.
}
    \begin{tabular}{ccccccc}
    \specialrule{.15em}{.05em}{.05em}
     OM. & EP. & BP. & KLD($\downarrow$) & SIM($\uparrow$) & NSS($\uparrow$) & IoU@0.4($\uparrow$) \\
    \hline
     \ccmark & \cxmark & \cxmark & 1.335 & 0.382 & 1.095 & 25.6 \\
     \ccmark & \ccmark & \cxmark & 1.193 & 0.439 & 1.278 & 32.5 \\
    \colortrainable \ccmark & \ccmark & \ccmark & \textbf{1.150} & \textbf{0.460} & \textbf{1.316} & \textbf{34.2} \\
    \specialrule{.15em}{.05em}{.05em}
    \end{tabular}
\label{tab:supp_ablation_mr_patch}
\end{table}

%% file: tabs/supp_ablation_backbone.tex
\begin{table*}[t]
\centering
\caption{\textbf{Ablation studies for the image encoder.}
We compare with the SOTA affordance grounding approach OOAL~\cite{li:ooal:2024}.
}
    \begin{tabular}{llcccc}
    \specialrule{.15em}{.05em}{.05em}
    Method & Backbone & KLD($\downarrow$) & SIM($\uparrow$) & NSS($\uparrow$) & IoU@0.4($\uparrow$) \\
    \hline
    & CLIP & 1.686 & 0.291 & 0.705 & 9.0 \\
    \multirow{-2}{*}{OOAL~\cite{li:ooal:2024}} & DINO & 1.652 & 0.314 & 0.753 & 9.4 \\
    \colortrainable & CLIP & 1.495 & 0.338 & 0.950 & 22.6 \\
    \colortrainable \multirow{-2}{*}{\ModelNametrain{}} & DINO & \textbf{1.374} & \textbf{0.368} & \textbf{1.053} & \textbf{22.9} \\
    \specialrule{.15em}{.05em}{.05em}
    \end{tabular}
\label{tab:supp_ablation_backbone}
\end{table*}

%% file: tabs/supp_ablation_text.tex
\begin{table}[t]
\centering
\caption{\textbf{Ablation studies for the text encoder.}
}
    \begin{tabular}{ccccc}
    \specialrule{.15em}{.05em}{.05em}
      Text encoder & KLD($\downarrow$) & SIM($\uparrow$) & NSS($\uparrow$) & IoU@0.4($\uparrow$) \\
    \hline
    \colortrainable ViT-B/32 & \textbf{1.150} & \textbf{0.460} & \textbf{1.316} & \textbf{34.2} \\
    ViT-g/14 & 1.152 & 0.445 & 1.306 & 33.4 \\
    \specialrule{.15em}{.05em}{.05em}
    \end{tabular}
\label{tab:supp_ablation_text}
\end{table}

%% file: tabs/supp_ablation_pl.tex
\begin{table}[t]
\centering
\caption{\textbf{Ablation studies for the prompt learning.}
`PL.' indicates whether the text prompt learner proposed by OOAL is applied.
}
    \begin{tabular}{lccccc}
    \specialrule{.15em}{.05em}{.05em}
    Model & PL. & KLD($\downarrow$) & SIM($\uparrow$) & NSS($\uparrow$) & IoU@0.4($\uparrow$) \\
    \hline
    OOAL & \ccmark & 1.281 & 0.425 & 1.152 & 19.8 \\
    OOAL & \cxmark & 1.253 & 0.437 & 1.163 & 21.1 \\
    Ours & \ccmark & 1.215 & 0.427 & 1.235 & 31.3 \\
    \colortrainable Ours & \cxmark & \textbf{1.150} & \textbf{0.460} & \textbf{1.316} & \textbf{34.2} \\
    \specialrule{.15em}{.05em}{.05em}
    \end{tabular}
\label{tab:supp_ablation_pl}
\end{table}

%% file: tabs/supp_hyper_postprocess.tex
\begin{table}[t]
  \centering
    \caption{\textbf{Hyperparameter search for the proposed patch refinement module in \ModelNamefree{}.}
    We show the hyperparameter search for the number of evenly-focused patches $e$ and border ratio $b$.
    }
      \begin{tabular}{cccc}
        \specialrule{.15em}{.05em}{.05em}
          \# of evenly-focused & \multicolumn{3}{c}{border ratio $b$} \\
          \cline{2-4}
          patches $e$ & 0.1 & 0.2 & 0.3 \\
          \hline
          4 & \textbf{20.5} & \textbf{20.5} & 20.2 \\
          9 & \textbf{20.5} & \cellcolor{trainfree}\textbf{20.5} & 20.2 \\
          16 & \textbf{20.5} & \textbf{20.5} & 20.2 \\
        \specialrule{.15em}{.05em}{.05em}
      \end{tabular}
  
  \label{tab:supp_hyper_postprocess}
\end{table}

%% file: tabs/supp_hyper_init.tex
\begin{table}[t]
\centering
\caption{\textbf{Hyperparameter search for the branch initialization when training \ModelNametrain{}.}
`Init. $\alpha$' represents the initialized value for $\alpha$ during training.
}
    \begin{tabular}{ccccc}
    \specialrule{.15em}{.05em}{.05em}
      Init. $\alpha$ & KLD($\downarrow$) & SIM($\uparrow$) & NSS($\uparrow$) & IoU@0.4($\uparrow$) \\
    \hline
    0 & 1.562 & 0.356 & 0.902 & 22.6  \\
    \colortrainable 0.5 & \textbf{1.150} & \textbf{0.460} & \textbf{1.316} & \textbf{34.2} \\
    1 & 1.171 & 0.450 & 1.246 & 33.5 \\
    \specialrule{.15em}{.05em}{.05em}
    \end{tabular}
\label{tab:supp_hyper_init}
\end{table}

%% file: tabs/supp_hyper_lr.tex
\begin{table}[t]
\centering
\caption{\textbf{Hyperparameter search for the learning rate when training \ModelNametrain{}.}
`Lr.' represents the learning rate.
}
    \begin{tabular}{ccccc}
    \specialrule{.15em}{.05em}{.05em}
      Lr. & KLD($\downarrow$) & SIM($\uparrow$) & NSS($\uparrow$) & IoU@0.4($\uparrow$) \\
    \hline
    1e-5 & 1.227 & 0.416 & 1.229 & 31.1 \\
    5e-5 & 1.242 & 0.406 & 1.233 & 31.3  \\
    \colortrainable 1e-4 & \textbf{1.150} & \textbf{0.460} & 1.316 & \textbf{34.2} \\
    5e-4 & 1.156 & 0.446 & \textbf{1.344} & 32.6  \\
    1e-3 & 1.174 & 0.434 & 1.265 & 32.4  \\
    \specialrule{.15em}{.05em}{.05em}
    \end{tabular}
\label{tab:supp_hyper_lr}
\end{table}

%% file: tabs/supp_computation.tex
\begin{table*}[t]
\centering
\caption{\textbf{Computational cost for the proposed \ModelNametrain{}.} We compare with the SOTA model OOAL~\cite{li:ooal:2024}.
}
    \begin{tabular}{lcccc}
    \specialrule{.15em}{.05em}{.05em}
     \multicolumn{1}{c}{} & \multicolumn{1}{c}{} & \multicolumn{3}{c}{Inference Time (s)} \\
     \cline{3-5}
     \multirow{-2}{*}{Method} & \multirow{-2}{*}{\# Trainable Parameters (M)} & Pre-processing & Inference & Total \\
    \hline
     OOAL~\cite{li:ooal:2024} & 
     \textbf{11.2} & - & \textbf{0.021} & \textbf{0.021} \\
    \colortrainable \ModelNametrain{} & 16.5 & 0.533 & 0.147 & 0.680\\
    \specialrule{.15em}{.05em}{.05em}
    \end{tabular}
\label{tab:supp_computation}
\end{table*}

%% file: figs/gpt5.tex
\begin{figure}[t]
\centering
\includegraphics[width=0.85\linewidth]{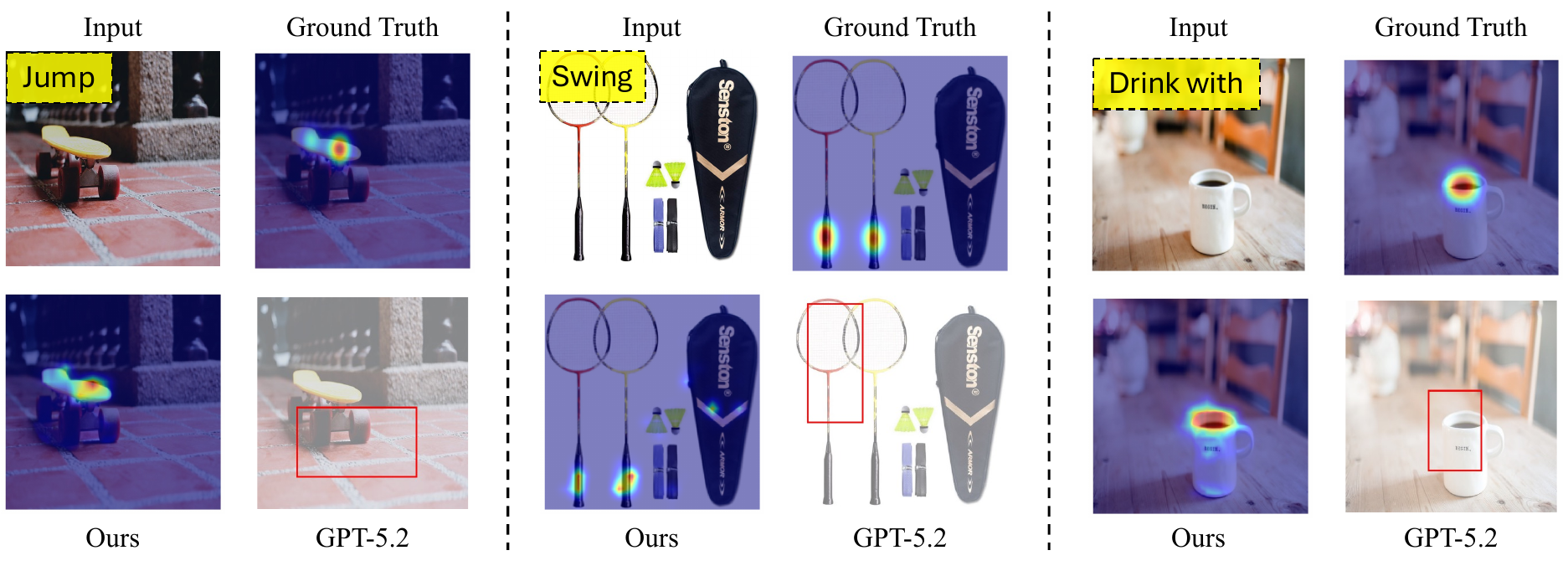}
\vspace{-0.3cm}
\caption{\textbf{Affordance grounding results with \texttt{GPT-5.2}.}
We prompt \texttt{GPT-5.2} to predict affordance bounding box coordinates in text form.
The predicted bounding boxes are overlaid with the original image for visualization. We observe that \texttt{GPT-5.2} is able to bound subparts of an object, however, the affordance regions are not reliably grounded.
}
\label{fig:gpt5}
\vspace{-0.5cm}
\end{figure}

%% file: figs/supp_qualitative.tex
\begin{figure*}[t]
\centering
\includegraphics[width=0.9\linewidth]{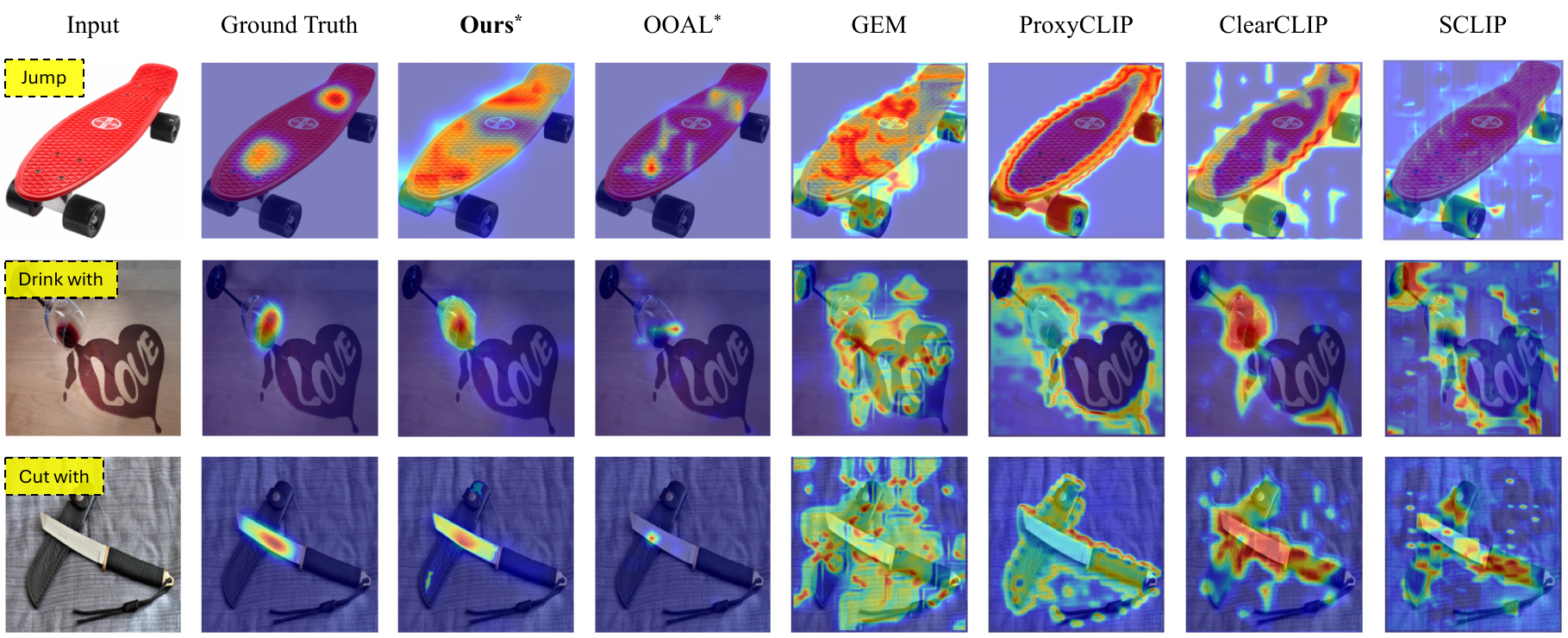}
\caption{\textbf{Additional qualitative results on the \agdbenchmark{} test set.}
$^*$ indicates that the model is trained with CLIP as the image encoder for fair comparison.
Predicted heatmaps are overlaid on the input image for visualization.}
\label{fig:supp_qualitative}
\end{figure*}

%% file: figs/supp_failure_cases.tex
\begin{figure}[t]
\centering
\includegraphics[width=0.6\linewidth]{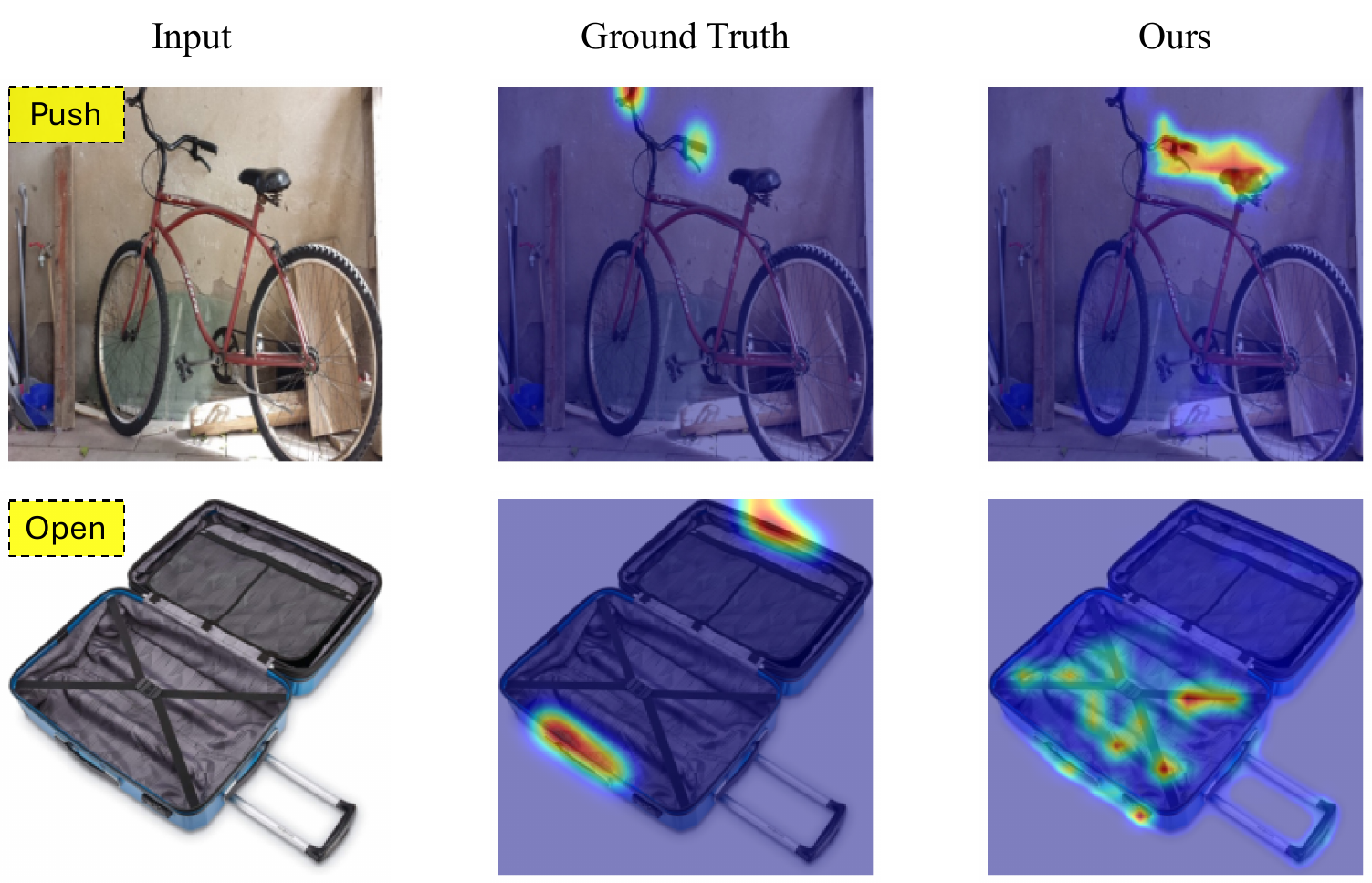}
\caption{\textbf{Failure cases on the \agdbenchmark{} test set.}
Predicted heatmaps are overlaid on the input image for visualization.}
\label{fig:supp_failure_cases}
\end{figure}

%% file: tabs/supp_add_datasplits.tex
\begin{table*}[t]
\centering
\caption{
\textbf{Additional data splits to verify the model's robustness.}
}
\resizebox{1\textwidth}{!}{
    \begin{tabular}{lccc}
    \specialrule{.15em}{.05em}{.05em}
        Split & Train (67 samples) & Val (253 samples) & Test (273 samples) \\
        \hline
        & take\_photo, throw, wash, peel, pour & push, drink\_with, brush\_with
        & text\_on, stick\\
        & sip, talk\_on, drag, pack, ride, sit\_on & boxing, catch, open, jump
        & cut, hit\\
        \multirow{-3}{*}{Additional split 1} & carry, eat, hold, kick, pick\_up & cut\_with, stir, swing, lift
        & lie\_on\\
        \hline
        & drag, pour, eat, talk\_on, write, type\_on, cut\_with, lift & drink\_with, lie\_on
        & sip, pick\_up \\
        & text\_on, hold, stick, open, kick, wash, throw, boxing & swing, brush\_with
        & peel, catch \\
        \multirow{-3}{*}{Additional split 2} & jump, beat, push, look\_out, take\_photo, sit\_on, cut & stir, pack, hit
        & carry, ride\\
    \specialrule{.15em}{.05em}{.05em}
    \end{tabular}
}
\label{tab:supp_add_datasplits}
\end{table*}

%% file: tabs/rel_work.tex
\begin{table}[t]
\centering
\caption{\textbf{Zero-shot settings of recent affordance grounding methods.}
Most existing approaches evaluate generalization to unseen \textit{object categories} (denoted by `Obj.').
In contrast, our method is the first to quantitatively evaluate generalization to novel \textit{affordance types} (denoted by `Aff.') in 2D.
\ccmark{} and \cxmark{} indicate the presence or absence of quantitative evaluation for the corresponding setting, respectively.
$^{\dag}$ indicates the method is evaluated on novel affordance types that are the synonyms of training affordances.
$^{\ddag}$ indicates the method requires additional auxiliary input during inference time.
}
    \begin{tabular}{lccc}
    \specialrule{.15em}{.05em}{.05em}
    & & \multicolumn{2}{c}{Zero-shot setting} \\
    \cline{3-4}
    \multirow{-2}{*}{Method} & \multirow{-2}{*}{Domain} & \makebox[3em]{Obj.} & \makebox[3em]{Aff.} \\
    \hline
    \rowcolor{lightgray} OpenAD$^{\dag}$~\cite{nguyen2023open} \textcolor{gray}{\textsubscript{[IROS'23]}} & 3D & \ccmark & \ccmark \\
    \rowcolor{lightgray} IAG~\cite{yang2023grounding} \textcolor{gray}{\textsubscript{[ICCV'23]}} & 3D & \ccmark & \cxmark \\
    \rowcolor{lightgray} LASO~\cite{li2024laso} \textcolor{gray}{\textsubscript{[CVPR'24]}} & 3D & \ccmark & \cxmark \\
    \rowcolor{lightgray} MIFAG~\cite{gao2025learning} \textcolor{gray}{\textsubscript{[AAAI'25]}} & 3D & \ccmark & \cxmark \\
    \rowcolor{lightgray} 3D-ADLLM$^{\dag}$~\cite{chu20253d} \textcolor{gray}{\textsubscript{[ICLR'25]}} & 3D & \ccmark & \ccmark \\
    \rowcolor{lightgray} GREAT$^{\ddag}$~\cite{shao2024great} \textcolor{gray}{\textsubscript{[CVPR'25]}} & 3D & \ccmark & \ccmark \\
    \rowcolor{lightgray} LMAffordance3D~\cite{zhu2025grounding} \textcolor{gray}{\textsubscript{[CVPR'25]}} & 3D & \ccmark & \cxmark \\
    \rowcolor{lightgray} GEAL~\cite{lu2025geal} \textcolor{gray}{\textsubscript{[CVPR'25]}} & 3D & \ccmark & \cxmark \\
    \rowcolor{lightgray} SeqAfford~\cite{yu2025seqafford} \textcolor{gray}{\textsubscript{[CVPR'25]}} & 3D & \ccmark & \cxmark \\
    \rowcolor{lightgray} 3DAffordSplat~\cite{wei20253daffordsplat} \textcolor{gray}{\textsubscript{[ACMMM'25]}} & 3D & \ccmark & \ccmark \\
    \rowcolor{lightgray} GLANCE~\cite{li2025intermediate} \textcolor{gray}{\textsubscript{[ICCV'25]}} & 3D & \ccmark & \cxmark \\
    \rowcolor{lightgray} ViSPLA~\cite{basakvispla2025} \textcolor{gray}{\textsubscript{[NeurIPS'25]}} & 3D & \ccmark & \cxmark \\
    
    \hline
    Cross-view-AG~\cite{Learningluo} \textcolor{gray}{\textsubscript{[CVPR'22]}} & 2D & \ccmark & \cxmark \\
    OSAD-Net~\cite{zhai2022one} \textcolor{gray}{\textsubscript{[IJCV'22]}} & 2D & \ccmark & \cxmark \\
    LOCATE~\cite{li2023locate} \textcolor{gray}{\textsubscript{[CVPR'23]}} & 2D & \ccmark & \cxmark \\
    WSMA~\cite{xu2024weakly} \textcolor{gray}{\textsubscript{[AAAI'24]}} & 2D & \ccmark & \cxmark \\
    OOAL~\cite{li:ooal:2024} \textcolor{gray}{\textsubscript{[CVPR'24]}} & 2D & \ccmark & \cxmark \\
    HOI-LOCATE~\cite{rai2024strategies} \textcolor{gray}{\textsubscript{[CVPRW'24]}} & 2D & \ccmark & \cxmark \\
    AffordanceLLM~\cite{qian2024affordancellm} \textcolor{gray}{\textsubscript{[CVPRW'24]}}& 2D & \ccmark & \cxmark \\
    AffordanceCLIP~\cite{cuttano2024does} \textcolor{gray}{\textsubscript{[CVPRW'24]}} & 2D & \ccmark & \cxmark \\
    OVAL-Prompt~\cite{tong2024oval} \textcolor{gray}{\textsubscript{[ICRAW'24]}} & 2D & \ccmark & \cxmark \\
    Robo-ABC~\cite{ju2024robo} \textcolor{gray}{\textsubscript{[ECCV'24]}} & 2D & \ccmark & \cxmark \\
    INTRA~\cite{jang2024intra} \textcolor{gray}{\textsubscript{[ECCV'24]}} & 2D & \ccmark & \cxmark \\
    WSAG-PLSP~\cite{xuweakly} \textcolor{gray}{\textsubscript{[ICLR'25]}} & 2D & \ccmark & \cxmark \\
    GAT~\cite{li2024learning} \textcolor{gray}{\textsubscript{[ICCV'25]}} & 2D & \ccmark & \cxmark \\
    SelectiveCL~\cite{moon2025selective} \textcolor{gray}{\textsubscript{[ICCV'25]}} & 2D & \ccmark & \cxmark \\
    LoopTrans~\cite{tang2025closed} \textcolor{gray}{\textsubscript{[ICCV'25]}} & 2D & \ccmark & \cxmark \\
    \colortrainable \textbf{Ours} & 2D & \ccmark & \ccmark \\
    \specialrule{.15em}{.05em}{.05em}
    \end{tabular}
\label{tab:rel_work}
\end{table}